\documentclass{article}

\PassOptionsToPackage{numbers, compress}{natbib}

\usepackage[eandd, final]{neurips_2026}
\usepackage[utf8]{inputenc}
\usepackage[T1]{fontenc}
\usepackage{hyperref}
\usepackage{booktabs}
\usepackage{amsfonts}
\usepackage{microtype}
\usepackage{amsmath}
\usepackage{multirow}
\usepackage{amssymb}
\usepackage{threeparttable}
\usepackage{makecell}
\usepackage{enumitem}
\usepackage{fontawesome5}
\usepackage{wrapfig}
\usepackage{caption}
\usepackage{xurl}
\usepackage{rotating}
\usepackage{float}

\usepackage[table]{xcolor}
\usepackage{graphicx}

\definecolor{row-green}{HTML}{F1F6EC}
\definecolor{row-yellow}{HTML}{FFFFF0}
\definecolor{row-pink}{HTML}{FFF5F5}
\definecolor{row-gray}{HTML}{E8E8E8}

\title{MySign: A High-Fidelity Motion-Capture Dataset for 3D Sign Generation in Bahasa Isyarat Malaysia}

\author{%
  Jiayu Shen$^{1,5}$ \quad
  Kalin Stefanov$^{2}$ \quad
  Lay-Ki Soon$^{1,5}$ \quad
  Vee Yee Chong$^{3}$ \\
  \bfseries
  Alpha Agape Gopalai$^{4}$ \quad
  KokSheik Wong$^{1,5}$ \\[0.5em]
  $^{1}$School of Information Technology, Monash University Malaysia \\
  $^{2}$Japan Advanced Institute of Science and Technology \\
  $^{3}$Jeffrey Cheah School of Medicine \& Health Sciences, Monash University Malaysia \\
  $^{4}$School of Engineering, Monash University Malaysia \\
  $^{5}$Action Lab Malaysia, Monash University Malaysia
}

\begin{document}

\maketitle

\begin{center}
  \includegraphics[width=0.85\linewidth, trim={0 0 0 0.5cm}, clip]{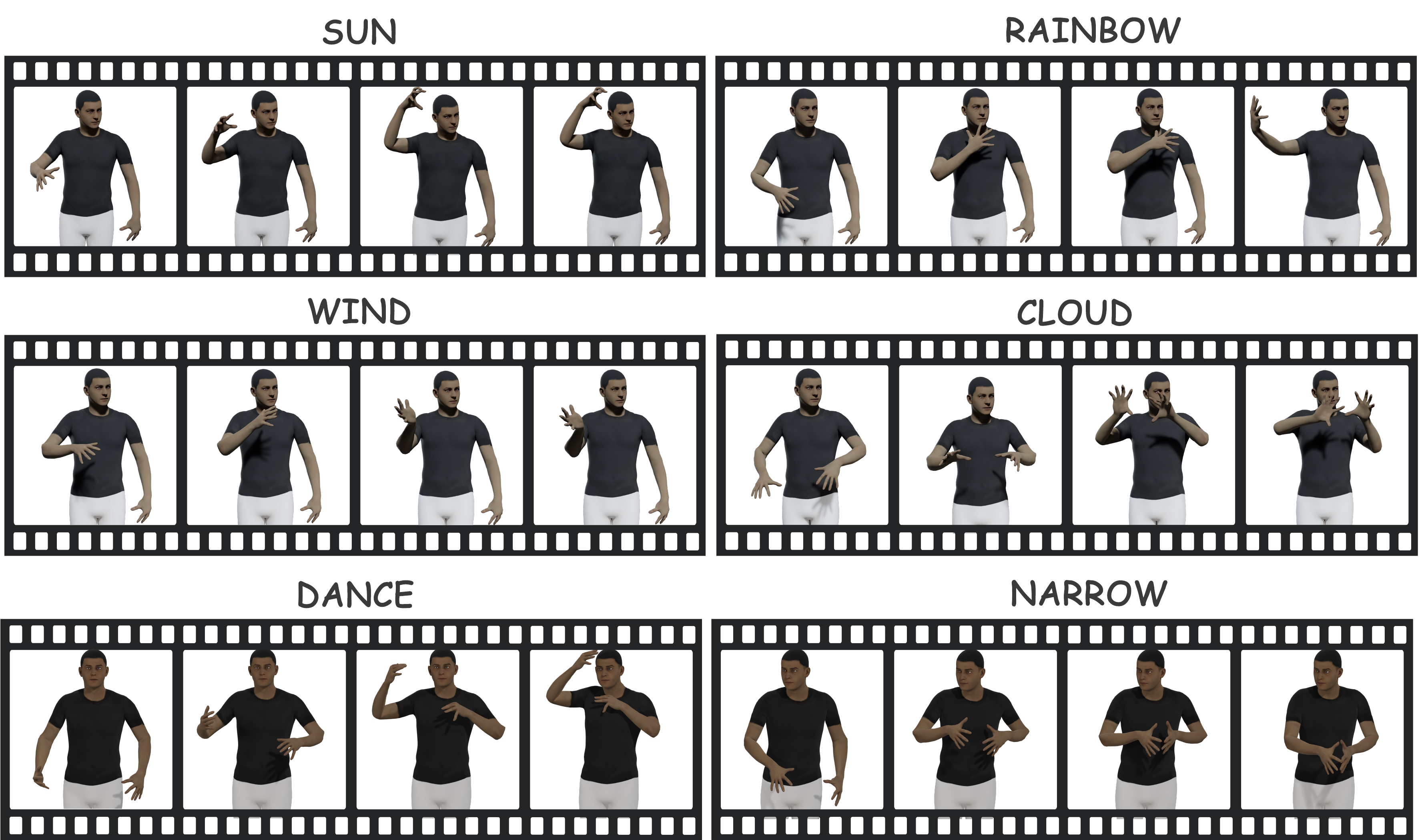}
  \captionof{figure}{Examples of isolated signs from MySign. We visualize four BIM signs as sequences of SMPL-X body and hand poses, illustrating the dataset's fine-grained hand articulation across time.}
  \label{fig:mysign}
\end{center}

\begin{abstract}
Progress in 3D sign generation is constrained by the scarcity of high-fidelity 3D sign language datasets, a limitation that is particularly severe for under-resourced languages.
We introduce MySign, the first motion-capture dataset for isolated signs in Bahasa Isyarat Malaysia (BIM).
Its 5,000 samples cover 1,000 lexical items (glosses) and were produced by five Deaf native signers.
Prior datasets that rely on SMPL-X parameters regressed from video provide pseudo-labels, which are often affected by depth ambiguity, inaccurate finger poses, and temporal jitter.
MySign instead provides ground-truth SMPL-X sequences retargeted from recordings made with a motion-capture setup tailored for sign language; these sequences offer higher-fidelity motion, particularly in fine-grained hand and finger articulation.
A controlled comparison on the same glosses and signers indicates that training on these sequences improves generation over training on such pseudo-labels.
We establish MySign as a challenging benchmark for gloss-conditioned 3D sign generation: state-of-the-art models produce visually plausible outputs but fall short of the ground truth, most notably in fine-grained articulation, revealing persistent gaps in modeling articulated sign motion.
In contrast, recognition baselines consistently achieve high lexical accuracy, indicating that the dataset reliably supports supervised learning.
The dataset is available at \url{https://huggingface.co/datasets/mysigner/MySign}.
\end{abstract}

\section{Introduction}
More than 70 million people worldwide rely on sign languages as their primary means of communication~\cite{deafculture1,deafculture2}.
Advances in AI have driven substantial progress in sign language recognition (SLR)~\cite{slr1,slr2,signbert} and sign language translation (SLT)~\cite{slt1,slt2,slt3}. In contrast, sign language generation (SLG)~\cite{pt,NSA} has received less attention, despite its critical role in hearing-to-deaf communication and accessibility applications.

Progress in SLR and SLT has been largely enabled by datasets focused on a few well-studied sign languages, including American Sign Language (ASL)~\cite{wlasl,how2sign,aslcitizen}, Australian Sign Language (Auslan)~\cite{MMWLAuslan,auslandaily}, Chinese Sign Language (CSL)~\cite{csl500,csldaily}, and German Sign Language (DGS)~\cite{phoenix14T}.
However, sign languages, like spoken languages, have their own grammars and lexicons and vary widely between regions~\cite{signgrammar1,manual1}, even where a common spoken language is shared.
For example, ASL and Auslan are linguistically distinct, differing in their phonological, lexical, and syntactic systems.
In Malaysia, the Malaysian Association of Sign Language Interpreters reported approximately 60 active interpreters serving more than 44,000 Deaf individuals in October 2025~\cite{chandran2025myasli_interpreters}---an accessibility gap that human interpretation alone cannot close.
Automated SLG provides a scalable path forward, but its development for low-resource sign languages such as Bahasa Isyarat Malaysia (BIM) remains limited by data scarcity: to our knowledge, no high-quality 3D dataset of isolated BIM signs is publicly available for SLG.

SLG aims to synthesize articulatorily accurate and realistic signing motions, which are primarily characterized by manual features (e.g., handshape, location, orientation, and movement)~\cite{manual1,manual2} and further enriched by non-manual features (e.g., facial expression, body orientation, and intensity).
However, as shown in Table~\ref{tab:sign_datasets}, most existing sign language datasets (e.g., WLASL~\cite{wlasl}, Phoenix-2014T~\cite{phoenix14T}, Auslan-Daily~\cite{auslandaily}) contain only RGB or RGB-D video. Without 3D pose annotations, they provide no direct supervision for fine-grained hand articulation and coordinated body motion, hindering generative models from learning accurate 3D sign representations.
Although 2D-to-3D pose lifting offers a hardware-free alternative to direct 3D motion capture (MoCap), its accuracy is inherently constrained because the task is ill-posed: depth ambiguity and occlusion often allow multiple plausible 3D poses to match a single 2D observation~\cite{2dto3d}.
Therefore, obtaining reliable 3D ground-truth annotations could be an important step toward more effective SLG modeling.
Recent efforts, such as 3D-LEX~\cite{3d-lex} and MC-TRISLAN~\cite{MCTRISLAN}, have begun to explore MoCap as a higher-fidelity acquisition modality.
Nevertheless, these datasets lack a unified parametric representation, limiting their utility for systematic benchmarking and fine-grained sign generation.
These limitations highlight a clear gap: the absence of an SLG dataset that combines high-fidelity 3D ground-truth annotations with a unified parametric representation suitable for modern generative modeling.

To address this gap, we present MySign, a sign language dataset for isolated BIM signs, comprising 5,000 high-fidelity samples performed by five Deaf native signers.
The dataset covers 1,000 unique glosses across nine semantic categories, including culture, daily life, health, and time.
The gloss vocabulary is curated from BIM Sign Bank, the official lexical resource for BIM~\cite{bimsignbank}.
MySign is captured using a multi-camera optical MoCap system and instrumented data gloves, yielding over 15.77 million synchronized pose frames with 3D skeletal annotations that can serve as ground truth for downstream sign language modeling tasks.
We further convert the raw recordings into a unified SMPL-X~\cite{smplx} representation with detailed hand and finger articulation (see Figure~\ref{fig:mysign}) through a fully reproducible post-processing pipeline, facilitating integration with modern generative models.
Designed for SLG research, MySign is, to our knowledge, the first MoCap dataset of isolated BIM signs.

To verify that MySign offers a reliable foundation, we first evaluate isolated sign recognition baselines.
Their high accuracy supports both the lexical discriminability of the gloss vocabulary and the fidelity of the SMPL-X annotations in preserving signing content.
Beyond recognition, our main benchmark for gloss-conditioned SLG shows that sign-specific methods outperform adapted general-purpose motion baselines, while still leaving a persistent gap from ground-truth motion.
This gap suggests that MySign captures linguistically meaningful finger articulation that current generative models have yet to reproduce faithfully.
To isolate the effect of data acquisition, we also train one sign-specific generator, at matched scale, on MoCap-grounded SMPL-X and on pseudo-labels estimated from RGB recordings of the same glosses by the same signers.
MoCap-grounded supervision yields lower FID and higher recognition accuracy than RGB-fitted supervision.
Overall, these findings establish MySign as both a high-quality training resource and an informative benchmark, underscoring fine-grained, linguistically faithful sign motion generation as an open challenge.
Our main contributions are as follows:
\begin{itemize}[leftmargin=*, noitemsep, topsep=0pt]
    \item Introducing MySign, the first publicly available MoCap dataset for isolated BIM sign generation, comprising 1,000 unique glosses across nine semantic categories and 5,000 samples performed by five Deaf native signers, totaling over 15.77M pose frames.
    \item Providing frame-aligned SMPL-X ground truth retargeted from synchronized multi-camera optical MoCap data and instrumented data-glove recordings captured in a calibrated studio. A controlled comparison indicates that these high-fidelity upper-body and hand kinematics provide better supervision for sign generation than RGB-fitted pseudo-labels.
    \item Benchmarking gloss-conditioned SLG on MySign with a predefined signer-independent split: sign-specific generators outperform adapted general-purpose motion baselines but still leave a persistent gap from ground-truth motion, while recognition baselines show that the motion encodes discriminative lexical information.
\end{itemize}
\section{Related Work}

\subsection{Sign Language Datasets}
\label{subsec:datasets}

\noindent\textbf{Video-based datasets.}
Large-scale sign language video datasets have driven substantial progress in SLR and SLT.
Phoenix-2014T~\cite{phoenix14T} and CSL-Daily~\cite{csldaily} are standard testbeds for continuous SLT in DGS and CSL, respectively, while WLASL~\cite{wlasl} serves as a large-vocabulary benchmark for isolated SLR in ASL.
These datasets primarily release near-frontal monocular RGB video paired with gloss and/or text annotations.
Methods requiring pose labels therefore estimate them from the video with off-the-shelf tools such as OpenPose~\cite{openpose} or MediaPipe~\cite{mediapipe}, adding monocular 2D-to-3D lifting~\cite{2dto3d} when 3D labels are needed.
The resulting estimates are 2D image-space keypoints or learned 3D approximations, leaving them vulnerable to depth ambiguity, self-occlusion, and the rapid handshape transitions characteristic of signing~\cite{SGNify,2dto3d2,2dto3d3}.

Multi-view datasets partially address these limitations: How2Sign~\cite{how2sign} covers ASL and includes a Panoptic Studio subset with 2D and 3D body and hand keypoints, and MM-WLAuslan~\cite{MMWLAuslan} extends such capture to Auslan.
These resources provide richer pose labels than monocular estimates, yet frequent hand--hand and hand--face occlusions still degrade their reliability when articulation matters most.
Since SLR and SLT are framed at the gloss or text level, the sparse, noisy keypoint or skeleton representations produced by RGB-based pipelines often suffice as discriminative features.
However, they lack the dense, mesh-level geometric and articulatory detail that high-fidelity 3D SLG requires.

\noindent\textbf{Reconstruction-based 3D datasets.}
To address this limitation, a parallel line of research has introduced 3D sign language motion datasets with holistic mesh annotations parameterized by SMPL-X~\cite{SGNify,NSA,dexavatar}.
SGNify~\cite{SGNify} established this paradigm, fitting SMPL-X to monocular videos under linguistic priors on hand poses.
NSA~\cite{NSA} scaled it to a full corpus, producing SMPL-X annotations for How2Sign with anatomical and temporal priors.
SignAvatars~\cite{signavatars} extended this paradigm to multiple sign languages, covering How2Sign, Phoenix-2014T, and several HamNoSys corpora, yielding the largest pseudo-labeled 3D sign dataset to date.
SOKE~\cite{soke} instead reused NSA's How2Sign annotations and re-fitted only CSL-Daily and Phoenix-2014T.
These pseudo-labeled datasets offer a practical way to scale 3D annotation to sign video corpora.
However, their SMPL-X annotations are fitted to monocular video under parametric priors rather than measured directly, and are thus affected by depth ambiguity, inaccurate hand poses under occlusion, and bias toward the priors.
These limitations preclude using such pseudo-labels as ground truth for high-fidelity generation and motivate sign datasets acquired through direct motion capture.

\noindent\textbf{MoCap-based 3D datasets.}
A smaller body of work measures sign motion directly.
MC-TRISLAN~\cite{MCTRISLAN} offers 18 hours of continuous Czech Sign Language captured with a marker-based Vicon system.
However, its seven-marker hand configuration, unlike its dense body marker set, provides limited resolution for fine-grained hand and finger articulation.
3D-LEX~\cite{3d-lex} combines Vicon body capture with StretchSense data gloves to record 1,000 isolated signs in ASL and Nederlandse Gebarentaal (NGT) from five signers.
Its single-exemplar-per-gloss design, however, limits coverage of signer-specific variation.
Crucially, neither dataset is released in an SMPL-X-compatible format, leaving a representation gap for end-to-end 3D SLG.
MySign instead provides ground-truth SMPL-X sequences from a motion-capture setup tailored for sign language, with each gloss performed by multiple Deaf native signers.

\newsavebox{\signtabbox}
\newcommand{\signdatasettabular}{%
\begin{tabular}{lclrrrcccc}
\toprule
\multirow{2}{*}{\textbf{Dataset}} & \multirow{2}{*}{\textbf{Year}} & \multirow{2}{*}{\textbf{Lang.}} &
\multicolumn{3}{c}{\textbf{Scale}} &
\textbf{Capture} & \textbf{Hand} & \textbf{Body} &
\multirow{2}{*}{\textbf{Task}} \\
\cmidrule(lr){4-6}
& & & \textbf{Gloss} & \textbf{Signer} & \textbf{Sample} & \textbf{Modality} & \textbf{Source} & \textbf{Model} & \\
\midrule
\rowcolor{row-gray}
\multicolumn{10}{c}{\textbf{Video-based}} \\
\midrule
\rowcolor{row-green}
CSL 500~\cite{csl500} & 2018 & CSL & 500 & 50 & 125K & RGB-D & D-skel & -- & SLR/SLT \\
\rowcolor{row-green}
Phoenix-2014T~\cite{phoenix14T} & 2018 & DGS & 1{,}066 & 9 & 8{,}257 & RGB & -- & -- & SLR/SLT \\
\rowcolor{row-green}
BosphorusSign22k~\cite{bosphorussign22k} & 2020 & T\.{I}D & 744 & 6 & 22{,}542 & RGB-D & D-skel & -- & SLR/SLT \\
\rowcolor{row-green}
AUTSL~\cite{autsl} & 2020 & T\.{I}D & 226 & 43 & 38{,}336 & RGB-D & D-skel & -- & SLR/SLT \\
\rowcolor{row-green}
WLASL~\cite{wlasl} & 2020 & ASL & 2{,}000 & 119 & 21{,}083 & RGB & RGB-kp & -- & SLR/SLT \\
\rowcolor{row-green}
How2Sign~\cite{how2sign} & 2021 & ASL & -- & 11 & 35{,}191 & RGB-D & RGB-kp & -- & SLR/SLT \\
\rowcolor{row-green}
CSL-Daily~\cite{csldaily} & 2021 & CSL & 2{,}000 & 10 & 20{,}654 & RGB & -- & -- & SLR/SLT \\
\rowcolor{row-green}
Auslan-Daily~\cite{auslandaily} & 2023 & Auslan & 600 & 67 & 25{,}106 & RGB & RGB-kp & -- & SLR/SLT \\
\rowcolor{row-green}
MM-WLAuslan~\cite{MMWLAuslan} & 2024 & Auslan & 3{,}215 & 73 & 282K+ & RGB-D & D-skel & -- & SLR/SLT \\
\midrule
\rowcolor{row-gray}
\multicolumn{10}{c}{\textbf{Reconstruction-based 3D}} \\
\midrule
\rowcolor{row-yellow}
SGNify~\cite{SGNify} & 2023 & DGS & 57 & 1 & 57 & RGB & Fit & SMPL-X & SLG \\
\rowcolor{row-yellow}
ASL3DWord~\cite{SignAvatar} & 2024 & ASL & 103 & -- & 1{,}547 & RGB & Fit & SMPL-X & SLG \\
\rowcolor{row-yellow}
NSA~\cite{NSA} & 2024 & ASL & -- & 11 & 35{,}191 & RGB & Fit & SMPL-X & SLG \\
\rowcolor{row-yellow}
SignAvatars~\cite{signavatars} & 2024 & 6 langs & -- & 153 & 70K & RGB & Fit & SMPL-X & SLG \\
\rowcolor{row-yellow}
SOKE~\cite{soke} & 2025 & CSL, DGS & -- & 19 & 28{,}911 & RGB & Fit & SMPL-X & SLG \\
\midrule
\rowcolor{row-gray}
\multicolumn{10}{c}{\textbf{MoCap-based 3D}} \\
\midrule
\rowcolor{row-pink}
MC-TRISLAN~\cite{MCTRISLAN} & 2022 & \v{C}ZJ & -- & 6 & -- & MoCap & Markers & -- & -- \\
\rowcolor{row-pink}
3D-LEX~\cite{3d-lex} & 2024 & ASL, NGT & 2$\times$1{,}000 & 5 & 2{,}000 & MoCap & Glove & -- & SLR/SLT \\
\rowcolor{row-pink}
\textbf{MySign} & \textbf{2026} & \textbf{BIM} & \textbf{1{,}000} & \textbf{5} & \textbf{5{,}000} & \textbf{MoCap} & \textbf{Glove} & \textbf{SMPL-X} & \textbf{SLG} \\
\bottomrule
\end{tabular}%
}

\begin{table}[t]
\captionsetup{position=top, skip=6pt} 
\caption{Comparison of sign language datasets, grouped by 3D signal acquisition. \emph{Hand Source} gives the origin of hand-articulation labels: \emph{Glove} or \emph{Markers} (direct capture), \emph{Fit} (parametric fitting to RGB), \emph{RGB-kp} (RGB-derived 2D/3D keypoints), or \emph{D-skel} (depth-sensor skeleton). T\.{I}D: Turkish Sign Language; \v{C}ZJ: Czech Sign Language; --: not available or not applicable.}
\label{tab:sign_datasets}
\centering
\tiny
\setlength{\tabcolsep}{0pt}%
\sbox{\signtabbox}{\signdatasettabular}%
\setlength{\tabcolsep}{\dimexpr(\textwidth-\wd\signtabbox)/20\relax}%
\signdatasettabular
\end{table}

\subsection{Human Motion and Sign Language Generation}

Human motion generation aims to synthesize realistic, temporally coherent body movements conditioned on different input modalities such as action labels, audio, or natural language~\cite{motionsurvey,actor,edge}.
Recent approaches typically encode motion as discrete latent tokens via autoencoders and model their distribution with autoregressive transformers~\cite{t2mgpt,momask,motiongpt}, or use diffusion models to generate dynamically plausible motion~\cite{mdm,mld,motiondiffuse}.
However, these methods mainly target full-body motion and largely overlook intricate hand and finger articulation~\cite{humanml3d,motionx}.
In contrast, sign language relies on fine-grained hand and upper-body movements that convey the primary linguistic meaning~\cite{signgrammar1}.
Prior work on sign language generation has focused on continuous signing, modeling sentence-level pose sequences and emphasizing global coherence, long-range dependencies, and smooth transitions between signs~\cite{pt,t2sgpt,continuousslg}.
We instead focus on isolated sign generation, which targets sign-internal articulation: the handshape, location, orientation, and movement that form an individual sign.
Contextual phenomena such as coarticulation are outside our scope.

Isolated sign generation remains challenging~\cite{wsigngen}: it requires fine-grained control over hand articulation and is further hindered by the scarcity of real-world 3D signing data.
Existing 3D datasets of continuous signing provide limited supervision for this articulation.
Their hand motion is typically either fitted to RGB video, yielding pseudo-labels prone to the errors discussed in Sec.~\ref{subsec:datasets}, or captured with sparse hand markers~\cite{MCTRISLAN}.
Moreover, these datasets are not segmented into isolated signs at capture time.
MySign addresses both limitations: data gloves directly measure finger motion, and each sequence is recorded for a single, predefined gloss (Sec.~\ref{subsec:recording}).
\section{MySign Dataset}
\label{sec:mysign}

\subsection{Participants and Ethics}
\label{subsec:participants}

\noindent\textbf{Participants statistics.}
The data were produced by five Deaf native signers (three male, two female) from four ethnic backgrounds in the Malaysian Deaf community: Malay, Chinese, Indian, and Kadazan.
The signers were 25 to 60 years old and 150 to 178\,cm tall.
Rather than maximizing the number of signers, we prioritize native fluency in BIM and have every signer cover the full vocabulary.
Small signer counts are common in sign language datasets (e.g., nine in Phoenix-2014T~\cite{phoenix14T} and eleven in How2Sign~\cite{how2sign}); for motion capture, the number of native signers available for studio sessions is the binding constraint.

\noindent\textbf{Ethics, consent, and compensation.}
All procedures were reviewed and approved by the Monash University Human Research Ethics Committee (MUHREC; Project ID 46717).
Recruitment proceeded through relevant mailing lists and snowball sampling within Deaf community networks.
To ensure accessibility, all recruitment and consent materials were delivered primarily in BIM, including short explanatory videos.
Each participant read and signed a consent form prior to their first session, and participation was entirely voluntary.
Participants received 50\,MYR per session, which exceeds the local minimum wage.

\subsection{Capture Setup and Recording Protocol}
\label{subsec:recording}
\noindent\textbf{Hardware and calibration.}
Recordings were conducted in a dedicated MoCap laboratory using the multimodal setup detailed in Appendix~\ref{app:capture-setup} (see Figure~\ref{fig:setup}).
We recorded upper-body kinematics using a six-camera OptiTrack system at 200\,Hz.
Each participant was fitted with 35 reflective markers placed according to a standard clinical protocol~\cite{isb}.
To address the limited resolution of optical markers on fine-grained per-finger articulation, signers wore MANUS Prime 3 Data gloves that capture finger motion directly.
The optical system and the data gloves were synchronized via a hardware trigger, producing a time-aligned stream of upper-body and finger kinematics.
Calibration was performed at two levels: 1) the gloves were calibrated in MANUS Core using a set of predefined hand poses to adapt the glove output to each signer's hand and 2) at the start of each session, an optical calibration was verified by confirming a wand calibration residual below 1\,mm and full marker visibility throughout the capture volume.
Signers wore everyday clothing fitted with small, unobtrusive markers to minimize interference with natural signing while retaining high-precision kinematic tracking.

\noindent\textbf{Recording protocol.}
Each session began with a setup phase in which the research team fitted the participant with the marker set and data gloves, followed by per-signer glove calibration and verification of the optical system.
An on-site coordinator then oversaw the start and end of recording.
Once setup was complete, signs were prompted via pre-designed instruction slides that displayed both the target gloss and a reference sign video, both taken directly from the authorized BIM Sign Bank~\cite{bimsignbank}.
Signs were recorded in a fixed category order to ensure coverage and reproducibility across participants.

\subsection{Data Processing and Annotation}

\noindent\textbf{Motion-capture processing.}
Marker trajectories and data-glove finger signals were processed with OptiTrack's Motive 3.0.3 software to reconstruct 3D joint positions and orientations, yielding a fused body-and-finger skeleton for each recording.

\noindent\textbf{SMPL-X retargeting.}
To obtain a unified representation compatible with existing motion generation models, we retarget the fused skeleton onto the canonical SMPL-X body model using the Rokoko Retarget plugin (v1.4.3) in Blender 4.5.2. Since the plugin performs direct bone mapping rather than model fitting, no fitting residual with respect to the raw MoCap and glove signals is available. Instead, we validate the retargeting by its kinematic consistency with the fused skeleton (Sec.~\ref{subsec:quality}).

\noindent\textbf{Gloss annotation.}
Each instance in MySign is recorded against a BIM Sign Bank prompt (Sec.~\ref{subsec:recording}), fixing its gloss at capture time rather than through post-hoc annotation.
By construction, every gloss label is taken directly from the Sign Bank lexicon, avoiding the inter-rater disagreement common in retrospectively labeled sign corpora.

\subsection{Quality Control and Validation}
\label{subsec:quality}

\noindent\textbf{Retargeting validation.}
We assess the kinematic consistency between the fused skeleton and the pre-cleaning SMPL-X output (i.e., before the ROM cleaning below) over the evaluated sequences (protocol in Appendix~\ref{app:rom:retargeting}).
Deviation is quantified by the bone-direction angle between corresponding bones and, for the PIP, DIP, and thumb joints, by the hand-joint position difference normalized by hand length.
Median bone-direction angles range from 3.0$^\circ$ at the shoulders to 7.0$^\circ$ at the thumb (95th percentiles 8.5$^\circ$--19.0$^\circ$; Table~\ref{tab:retargeting}), and the normalized hand-joint position difference has a median of 4.5\% (95th percentile 13.0\%).
Retargeting thus largely preserves the fused skeleton's bone directions and intra-hand configuration, with larger deviations at the distal finger and thumb joints.

\noindent\textbf{Anatomical range-of-motion cleaning.}
Although the motion-capture source is physically grounded, retargeting can introduce anatomically implausible SMPL-X joint rotations, most often over-flexed or over-abducted fingers.
We therefore clamp each joint rotation to its anatomical range of motion (ROM)~\cite{hmg1,hmg2}.
For joint $j$ with axis-angle rotation $\boldsymbol{\theta}_j \in \mathbb{R}^3$ and $\mathbf{R}_j = \exp([\boldsymbol{\theta}_j]_\times) \in \mathrm{SO}(3)$, we decompose $\mathbf{R}_j$ into Euler angles, clamp each component to a per-joint interval $[\mathbf{e}_j^{\min}, \mathbf{e}_j^{\max}]$, and recompose the rotation; this is an axis-aligned approximation to projection onto the ROM region.
For the shoulders, elbows, and wrists, clamping is applied directly in the native SMPL-X joint frames.
For finger joints, whose SMPL-X frames are not aligned with the flexion--abduction axes of clinical ROM, a per-joint change of basis $\mathbf{B}_j$ maps each rotation into an anatomical frame, where it is clamped before being mapped back.
The full configuration is described in Appendix~\ref{app:rom:config}.

Clip rates (Table~\ref{tab:rom_clamping}) remain below 6\% for the arm joints but rise along the finger chain (7.36\%/13.12\%/21.04\% at MCP/PIP/DIP), consistent with the growing retargeting deviation (median bone-direction angle 4.0$^\circ$/5.0$^\circ$/6.5$^\circ$).
To test whether hand clamping changes lexically meaningful handshapes, we run a paired ablation with and without it (Appendix~\ref{app:rom:clamping}).
Among clipped frames, the median fingertip displacement is only 0.15\,mm, and the frozen ST-GCN evaluator (Sec.~\ref{subsec:recognition}) predicts the same gloss in both conditions for 96.8\% of test sequences and for 92.0\% of the top decile by fingertip displacement.
Clamping thus largely preserves the recognized gloss, including on the most strongly corrected sequences.

\begin{table}[t]
  \captionsetup{position=top, skip=6pt}
  \caption{Per-joint-group ROM clamping statistics. \emph{Clip rate}: percentage of frames in which at least one Euler component of any joint in the group is actively clipped. \emph{Mean corr.}: mean magnitude of the clipping correction over the affected components.}
  \label{tab:rom_clamping}
  \centering
  \tiny
  \setlength{\tabcolsep}{4pt}
    \begin{tabular}{@{}llccc@{}}
      \toprule
      \textbf{Category} & \textbf{Joint group} & \textbf{\#\,joints} & \textbf{Clip rate (\%)} & \textbf{Mean corr.\ (\textdegree)} \\
      \midrule
      \multirow{3}{*}{Arm}
      & Shoulders (L/R)       &  2 &  3.11 & 2.87 \\
      & Elbows (L/R)          &  2 &  0.70 & 0.63 \\
      & Wrists (L/R)          &  2 &  5.89 & 2.43 \\
      \midrule
      \multirow{4}{*}{Hand}
      & Finger MCP (proximal) & 8 &  7.36 & 3.21 \\
      & Finger PIP (middle)   & 8 & 13.12 & 3.75 \\
      & Finger DIP (distal)   & 8 & 21.04 & 6.08 \\
      & Thumb joints (L/R)    &  6 & 16.98 & 5.19 \\
      \midrule
      \multicolumn{2}{@{}c}{\textbf{Joint-weighted mean}} & 36 & 12.60 & 4.09 \\
      \bottomrule
    \end{tabular}
\end{table}

\noindent\textbf{Expert-based perceptual validation.}
To characterize the recognizability and naturalness of the released sequences, we conducted a post-hoc expert review whose judgments serve as quality metadata and were not used to filter or modify the dataset.
Three Deaf native signers with BIM domain expertise participated.
For each instance, raters viewed only the rendered 3D sign through an interactive web interface, without access to the reference gloss, reported the perceived meaning, and rated motion naturalness on a
five-point scale (Very low to Very high).
Mean gloss-recognition accuracy was $86.3\%$ with a substantial inter-rater agreement (Fleiss' $\kappa = 0.73$), and the mean naturalness rating was $3.87 \pm 0.71$.
Full protocol details are provided in Appendix~\ref{app:verify}.

\subsection{Dataset Statistics and Splits}
\label{subsec:splits}

\begin{wrapfigure}{r}{0.61\textwidth}
  \centering
  \includegraphics[trim=30 20 150 32, clip, width=0.5\textwidth]{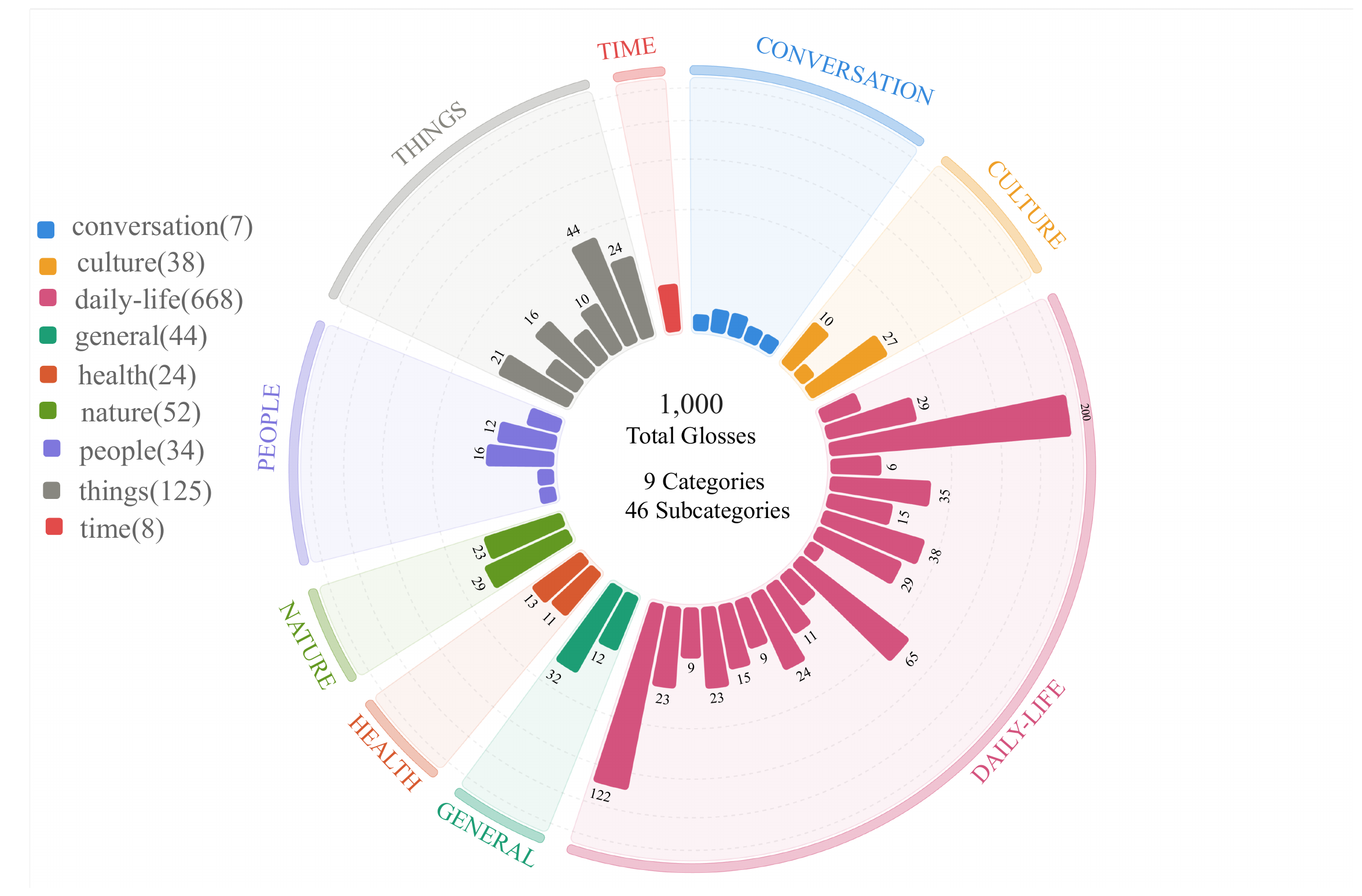}
  \caption{Overview of the lexical composition of MySign.}
  \label{fig:vocab_distribution}
\end{wrapfigure}

\noindent\textbf{Vocabulary construction.} MySign\footnote{
MySign is released at \url{https://huggingface.co/datasets/mysigner/MySign} under the Creative Commons Attribution-NonCommercial-ShareAlike 4.0 International license (CC BY-NC-SA 4.0).
All signers gave written informed consent prior to recording (Sec.~\ref{subsec:participants}).
} contains 1{,}000 BIM glosses selected from the BIM Sign Bank.
The research team made the selection under the guidance of a Deaf supervisor, prioritizing natural BIM signs commonly used in daily life and conversations so that the vocabulary reflects real communication scenarios.
To ensure diversity, the glosses were chosen to span nine categories and 46 subcategories, as shown in Figure~\ref{fig:vocab_distribution}.
The nine categories cover common conversations, daily activities, people, objects, time expressions, health-related communication, cultural concepts, nature-related terms, and other general concepts.

\noindent\textbf{Recording statistics.}
Each gloss is recorded once by each of five signers, forming a fully balanced gloss $\times$ signer composition with no missing entries.
The dataset comprises 5{,}000 isolated sign instances, 15.77M synchronized frames captured at 200\,Hz, and approximately 36 hours of recording.
The captured motions exhibit temporally continuous, finger-specific articulation and broad bimanual hand-pose variation.
Supporting visualizations and analysis are provided in Appendix~\ref{app:motion_visualization}.

\noindent\textbf{Splits.}
To ensure reproducibility, we provide a predefined signer-independent split with the dataset.
Of the five signers, four (S1--S4) are assigned to the training set and one (S5) to the test set, yielding a 4:1 ratio with no signer overlap.
All experiments on MySign, including those in Table~\ref{tab:crossdataset}, use this split; the only exception is the five-fold leave-one-signer-out evaluation in Table~\ref{tab:loso}, which holds out each signer in turn.
\section{Experiments}
\label{Sec:exp}

\begin{wraptable}{r}{0.38\textwidth}
\vspace{-4mm}
\centering
\footnotesize
\caption{Isolated sign language recognition accuracy (\%) of representative baselines on MySign, averaged over 20 evaluation repetitions.}
\label{tab:recognition}
\setlength{\tabcolsep}{4pt}
\begin{tabular}{lcc}
    \toprule
    \textbf{Method} & \textbf{Top-1} & \textbf{Top-5} \\
    \midrule
    ST-GCN~\cite{stgcn}           & 85.70 & 92.76 \\
    SignBERT~\cite{signbert}      & 87.33 & 94.35 \\
    BEST~\cite{best}              & 90.25 & 97.33 \\
    SignBERT+~\cite{signbert+}    & \textbf{92.90} & \textbf{98.48} \\
    \bottomrule
\end{tabular}
\vspace{-3mm}
\end{wraptable}

\subsection{Recognition-Based Validation of Lexical Content}
\label{subsec:recognition}
To evaluate whether MySign preserves lexical content, we train four representative SLR baselines---ST-GCN~\cite{stgcn}, SignBERT~\cite{signbert}, BEST~\cite{best}, and SignBERT+~\cite{signbert+}---and measure gloss-level classification accuracy.
We adapt all baselines to take SMPL-X joint rotations in the continuous 6D representation~\cite{6Dcontinuity} as input and report Top-1 and Top-5 accuracies.
As shown in Table~\ref{tab:recognition}, even the weakest baseline, ST-GCN, reaches 85.70\% Top-1 accuracy on the 1,000-way signer-independent test set, and the strongest, SignBERT+, reaches 92.90\%.
Consistently high accuracy across all baselines indicates that the 3D motion in MySign encodes discriminative lexical information and that the dataset reliably supports supervised learning.
For the downstream generation benchmark, we adopt the ST-GCN recognizer trained here as the default semantic evaluator, following SignAvatar~\cite{SignAvatar}.

\subsection{Gloss-Conditioned Isolated Sign Language Generation}

\noindent\textbf{Task definition.}
Given a single gloss $g \in \mathcal{V}$ drawn from a finite vocabulary $\mathcal{V}$, the goal is to generate a 3D upper-body motion sequence $\hat{M} = [\hat{p}_1, \dots, \hat{p}_T] \in \mathbb{R}^{T \times D}$ that conveys the lexical meaning of $g$.
Each frame $\hat{p}_t \in \mathbb{R}^{D}$ encodes the rotations of the upper-body SMPL-X joints (including the two hands) in the 6D rotation representation~\cite{6Dcontinuity}, and the sequence length $T$ is predicted per sample. 
Formally, we learn a conditional generator $f_\phi$ that, given $g$ and random noise $z$, produces $\hat{M} = f_\phi(g, z)$ whose distribution approximates the ground-truth distribution $p(M \mid g)$.
In our benchmark, $\mathcal{V}$ is the $1{,}000$-gloss vocabulary of MySign.

\begin{table}[t]
\centering
\footnotesize
\captionsetup{position=top, skip=6pt}
\caption{Gloss-conditioned isolated sign generation on MySign. $\uparrow$ / $\downarrow$ denote that higher / lower is better; $\rightarrow$ denotes that values closer to the \emph{Real} reference are preferred, and $\pm$ indicates the 95\% confidence interval.}
\label{tab:generation}
\setlength{\tabcolsep}{6pt}
\begin{tabular}{lcccc}
    \toprule
    \textbf{Method} & \textbf{FID$\downarrow$} & \textbf{Acc$\uparrow$} & \textbf{Div$\rightarrow$} & \textbf{Multimod$\rightarrow$} \\
    \midrule
    Real & 0.023$^{\scriptstyle \pm .020}$ & 0.857$^{\scriptstyle \pm .002}$ & 46.881$^{\scriptstyle \pm .300}$ & 18.719$^{\scriptstyle \pm .250}$ \\
    \midrule
    ACTOR~\cite{actor} & 143.293$^{\scriptstyle \pm 5.073}$ & 0.610$^{\scriptstyle \pm .005}$ & 36.472$^{\scriptstyle \pm .590}$ & 32.991$^{\scriptstyle \pm .320}$ \\
    MDM~\cite{mdm} & 119.367$^{\scriptstyle \pm 5.802}$ & 0.632$^{\scriptstyle \pm .007}$ & 35.621$^{\scriptstyle \pm .600}$ & 32.862$^{\scriptstyle \pm .380}$ \\
    MLD~\cite{mld} & 92.011$^{\scriptstyle \pm 4.241}$ & 0.626$^{\scriptstyle \pm .009}$ & 36.483$^{\scriptstyle \pm .600}$ & 36.824$^{\scriptstyle \pm .420}$ \\
    MotionDiffuse~\cite{motiondiffuse} & 83.564$^{\scriptstyle \pm 4.310}$ & 0.639$^{\scriptstyle \pm .007}$ & 36.538$^{\scriptstyle \pm .480}$ & 32.076$^{\scriptstyle \pm .340}$ \\
    LS-GAN~\cite{Ls-gan} & 131.331$^{\scriptstyle \pm 5.106}$ & 0.605$^{\scriptstyle \pm .011}$ & 36.531$^{\scriptstyle \pm .420}$ & 33.718$^{\scriptstyle \pm .340}$ \\
    \midrule
    SignAvatar~\cite{SignAvatar} & 54.166$^{\scriptstyle \pm 2.010}$ & 0.727$^{\scriptstyle \pm .021}$ & \textbf{48.299}$^{\scriptstyle \pm .730}$ & 22.913$^{\scriptstyle \pm .200}$ \\
    wSignGen~\cite{wsigngen} & \textbf{41.465}$^{\scriptstyle \pm 2.069}$ & \textbf{0.758}$^{\scriptstyle \pm .013}$ & 49.407$^{\scriptstyle \pm .810}$ & \textbf{22.825}$^{\scriptstyle \pm .230}$ \\
    \bottomrule
\end{tabular}
\end{table}

\noindent\textbf{Evaluation metrics.}
We adopt established evaluation metrics~\cite{SignAvatar}. 
Fr\'echet Inception Distance (FID) measures the distributional gap between generated and real signs.
Recognition accuracy (Acc), computed with the ST-GCN evaluator trained on MySign (Sec.~\ref{subsec:recognition}), evaluates whether generated motions can be classified as their target glosses.
Diversity (Div) captures feature-level variation across the full set of generated samples.
MultiModality (Multimod) captures variation among samples generated from the same gloss.
For each method, we report the mean and 95\% confidence interval over 20 runs of generation and evaluation with a single trained model.

\noindent\textbf{Baseline methods.}
We benchmark MySign with two groups of baselines.
The first group includes SignAvatar~\cite{SignAvatar} and wSignGen~\cite{wsigngen}, which, to the best of our knowledge, are the only publicly available methods specifically designed for isolated sign language generation.
The second group consists of general action-conditioned human motion generation methods, namely ACTOR~\cite{actor}, MDM~\cite{mdm}, MLD~\cite{mld}, MotionDiffuse~\cite{motiondiffuse}, and LS-GAN~\cite{Ls-gan}.
We adapt them to gloss conditioning by replacing the one-hot action label with the CLIP~\cite{clip} embedding of the gloss text, leaving their architectures otherwise unchanged.
For more experimental details, refer to Appendix~\ref{app:experimental_settings}.

\noindent\textbf{Benchmarking MySign.}
As shown in Table~\ref{tab:generation}, sign-specific methods substantially outperform the adapted general motion baselines on MySign.
wSignGen~\cite{wsigngen} achieves an FID of 41.5 and an Acc of 75.8\%, compared with 83.6 and 63.9\% for MotionDiffuse~\cite{motiondiffuse}, the best-performing general baseline.
This suggests that the general baselines struggle to capture the fine-grained articulation required for sign production.
Beyond fidelity, the two categories also diverge in distributional spread: sign-specific methods stay much closer to Real (Div 48.3--49.4 vs.\ 46.9; Multimod 22.8--22.9 vs.\ 18.7) than the general baselines (Div 35.6--36.5; Multimod 32.1--36.8).
For the general baselines, Multimod nearly matches Div, i.e., samples of the same gloss differ almost as much as samples of different glosses, which suggests weak gloss conditioning and is consistent with their lower Acc.
Nevertheless, a clear gap to Real remains on FID and Acc even for the best-performing sign-specific method, leaving considerable room for further advances in gloss-conditioned isolated sign generation on MySign. 

\noindent\textbf{Leave-one-signer-out evaluation.}
The predefined split holds out a single signer, so its results may depend on which signer is held out.
To investigate this, we repeat the evaluation with each of the five signers held out in turn (Table~\ref{tab:loso}).
In each fold, we report the recognition accuracy of ST-GCN, which also serves as that fold's evaluator, and compare wSignGen and MotionDiffuse, the best-performing sign-specific method and adapted general motion baseline, respectively.
Across folds, ST-GCN Top-1 accuracy ranges from 83.64\% to 88.39\%, and wSignGen achieves lower FID and higher Acc than MotionDiffuse in every fold.
Both the recognition accuracy and the ranking of the generation methods are thus robust to the choice of held-out signer.

\begin{table}[t]
\centering
\footnotesize
\captionsetup{position=top, skip=6pt}
\caption{Five-fold leave-one-signer-out evaluation. $\pm$ denotes the 95\% CI. S5 corresponds to the predefined signer-independent split in Table~\ref{tab:recognition}.}
\label{tab:loso}
\setlength{\tabcolsep}{6pt}
\begin{tabular}{lccccc}
    \toprule
    \multirow{2}{*}{\textbf{Held-out signer}} & \textbf{ST-GCN} & \multicolumn{2}{c}{\textbf{FID$\downarrow$}} & \multicolumn{2}{c}{\textbf{Acc$\uparrow$}} \\
    \cmidrule(lr){3-4} \cmidrule(lr){5-6}
     & \textbf{Top-1 (\%)} & \textbf{wSignGen} & \textbf{MotionDiffuse} & \textbf{wSignGen} & \textbf{MotionDiffuse} \\
    \midrule
    S1 & 87.42 & 44.018$^{\scriptstyle \pm 4.370}$ & 87.914$^{\scriptstyle \pm 6.075}$ & 0.771$^{\scriptstyle \pm .027}$ & 0.649$^{\scriptstyle \pm .044}$ \\
    S2 & 84.18 & 42.698$^{\scriptstyle \pm 2.072}$ & 85.453$^{\scriptstyle \pm 1.540}$ & 0.747$^{\scriptstyle \pm .011}$ & 0.622$^{\scriptstyle \pm .005}$ \\
    S3 & 88.39 & 36.821$^{\scriptstyle \pm 1.480}$ & 76.530$^{\scriptstyle \pm 3.061}$ & 0.784$^{\scriptstyle \pm .023}$ & 0.669$^{\scriptstyle \pm .008}$ \\
    S4 & 83.64 & 47.556$^{\scriptstyle \pm 1.991}$ & 98.611$^{\scriptstyle \pm 4.525}$ & 0.736$^{\scriptstyle \pm .019}$ & 0.611$^{\scriptstyle \pm .011}$ \\
    S5 & 85.70 & 41.465$^{\scriptstyle \pm 2.069}$ & 83.564$^{\scriptstyle \pm 4.310}$ & 0.758$^{\scriptstyle \pm .013}$ & 0.639$^{\scriptstyle \pm .007}$ \\
    \bottomrule
\end{tabular}
\end{table}

\noindent\textbf{MoCap-grounded vs.\ RGB-fitted supervision.}
To isolate the effect of data acquisition, we train wSignGen~\cite{wsigngen} under two supervision conditions matched in sample scale: MoCap-grounded and RGB-fitted.
For the latter, SMPL-X pseudo-labels are estimated with Hand4Whole~\cite{hand4whole} from separate RGB recordings of the same BIM glosses by the same signers; these recordings are used only for this comparison and are not part of the MySign release.
Both conditions are evaluated with a single frozen evaluator trained on pooled MoCap-real and RGB-real motions; all other settings are identical.
MoCap-grounded supervision yields higher Acc and lower FID against both references, including RGB-real, which shares the domain of the RGB-fitted training data (Table~\ref{tab:mocap_vs_rgb}), suggesting that the advantage stems from data acquisition rather than reference-domain matching.

\begin{table}[t]
\centering
\footnotesize
\captionsetup{position=top, skip=6pt}
\caption{MoCap-grounded vs.\ RGB-fitted supervision for wSignGen~\cite{wsigngen}. MoCap-real/RGB-real: real MoCap-grounded/RGB-fitted SMPL-X motions used as FID references.}
\label{tab:mocap_vs_rgb}
\setlength{\tabcolsep}{6pt}
\begin{tabular}{lccc}
    \toprule
    \textbf{Supervision} & \textbf{FID to MoCap-real$\downarrow$} & \textbf{FID to RGB-real$\downarrow$} & \textbf{Acc$\uparrow$} \\
    \midrule
    MoCap-grounded & 42.800$^{\scriptstyle \pm 2.200}$ & 45.600$^{\scriptstyle \pm 2.400}$ & 0.747$^{\scriptstyle \pm .012}$ \\
    RGB-fitted     & 51.400$^{\scriptstyle \pm 2.700}$ & 49.300$^{\scriptstyle \pm 2.600}$ & 0.704$^{\scriptstyle \pm .014}$ \\
    \bottomrule
\end{tabular}
\end{table}

\noindent\textbf{Comparison with ASL3DWord.}
We further compare MySign with ASL3DWord~\cite{SignAvatar}, the other publicly available 3D isolated sign language dataset in SMPL-X format, whose parameters are fitted from video.
This comparison is not a cross-lingual evaluation; rather, it tests whether pretraining on one dataset provides useful motion-level priors for learning the other's lexicon.
For each target dataset, we train SignAvatar~\cite{SignAvatar} on its training split either directly or after pretraining on the other dataset; in the latter case, we reinitialize the gloss-conditioning layer and fine-tune the full model.

Table~\ref{tab:crossdataset} shows an asymmetric transfer.
On ASL3DWord, MySign pretraining lowers FID relative to ASL3DWord-only training (63.831 vs.\ 71.809), with comparable Accuracy (0.751 vs.\ 0.733).
In the reverse direction, ASL3DWord pretraining degrades results on MySign, increasing FID (65.037 vs.\ 54.166) and lowering Accuracy (0.581 vs.\ 0.727) relative to MySign-only training.
This asymmetry is consistent with reusable motion-level priors learned from MySign rather than with cross-lingual lexical transfer.

\begin{figure}[h]
  \centering
  \includegraphics[width=0.85\linewidth, trim=20pt 45pt 10pt 10pt, clip]{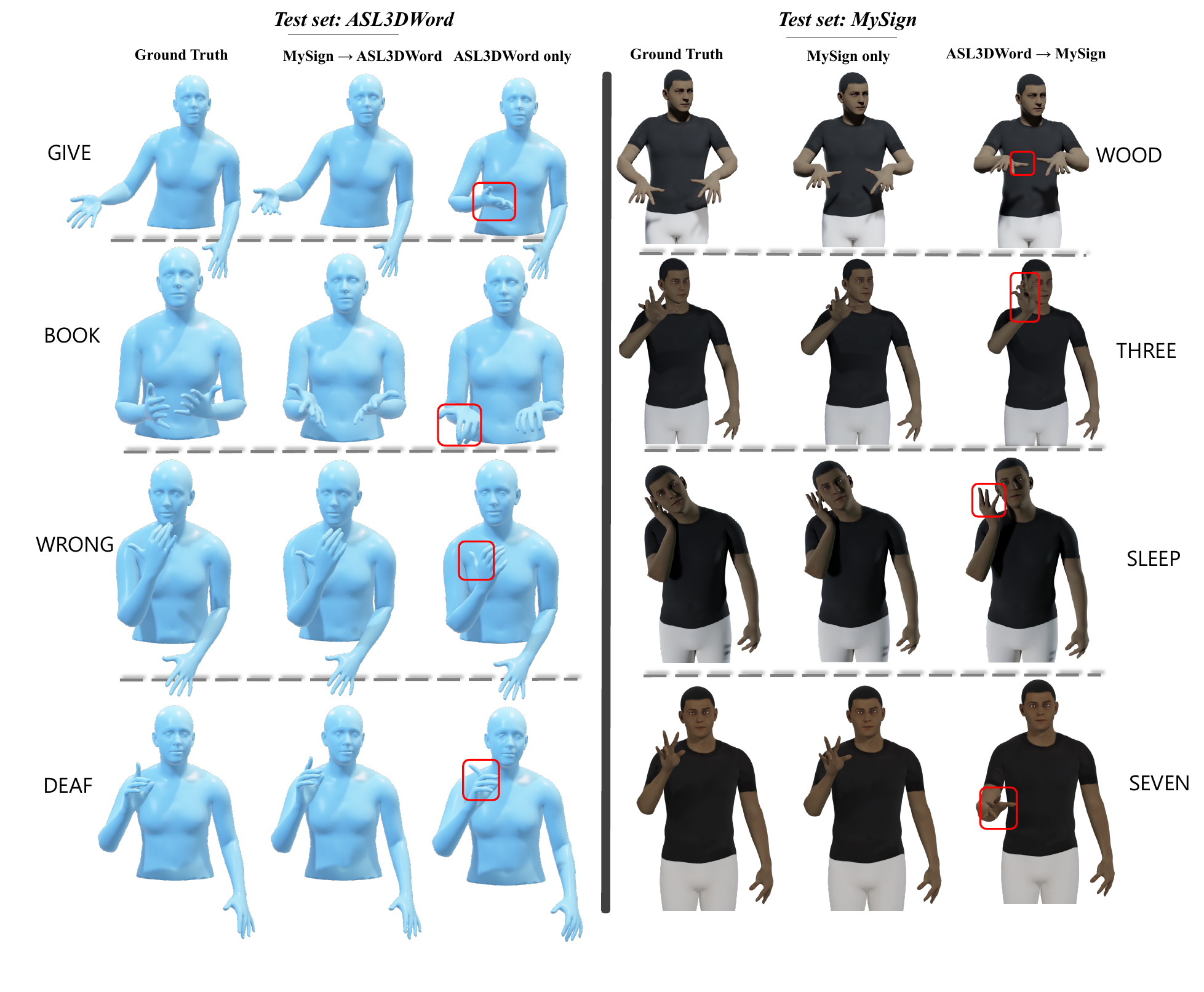}
  \caption{Qualitative comparison on the ASL3DWord (left) and MySign (right) test sets. $A \rightarrow B$: pretrained on $A$, fine-tuned on $B$. The panels contain different glosses and are compared only within each panel. The MySign$\rightarrow$ASL3DWord (left) and MySign-only (right) models produce hand configurations closer to the ground truth and fewer visible finger artifacts than their counterparts, as highlighted in the boxes.}
  \label{fig:qualitative}
\end{figure}

Figure~\ref{fig:qualitative} shows qualitative results on signs whose active handshape remains largely unchanged, which facilitates inspection of handshape fidelity and finger artifacts.
The two panels contain different glosses and are compared only within each panel.
In both panels, the model trained with MySign data (MySign-pretrained on the left, MySign-only on the right) better preserves plausible hand configurations than its counterpart, which exhibits noticeable finger artifacts and collapsed finger poses.
Together, these results indicate that the high-fidelity 3D motion in MySign benefits downstream sign generation, particularly in hand articulation.

\begin{table}[h]
\centering
\tiny
\captionsetup{position=top, skip=6pt}
\caption{Comparison between MySign and ASL3DWord using SignAvatar~\cite{SignAvatar}. Each column group is a target dataset, evaluated with its own recognizer; metrics are comparable only within a group. }
\label{tab:crossdataset}
\setlength{\tabcolsep}{4pt}
\begin{tabular}{l|cccc|cccc}
    \toprule
    \multirow{2}{*}{\textbf{Source}}
    & \multicolumn{4}{c|}{\textbf{ASL3DWord~\cite{SignAvatar} (Test)}}
    & \multicolumn{4}{c}{\textbf{MySign (Test)}} \\
    \cmidrule(lr){2-5} \cmidrule(lr){6-9}
    & \textbf{FID$\downarrow$} & \textbf{Acc$\uparrow$} & \textbf{Div$\rightarrow$} & \textbf{Multimod$\rightarrow$}
    & \textbf{FID$\downarrow$} & \textbf{Acc$\uparrow$} & \textbf{Div$\rightarrow$} & \textbf{Multimod$\rightarrow$} \\
    \midrule
    Real
    & 0.100$^{\scriptstyle \pm .010}$ & 0.818$^{\scriptstyle \pm .005}$ & 30.599$^{\scriptstyle \pm .450}$ & 12.289$^{\scriptstyle \pm .200}$
    & 0.023$^{\scriptstyle \pm .020}$ & 0.857$^{\scriptstyle \pm .002}$ & 46.881$^{\scriptstyle \pm .300}$ & 18.719$^{\scriptstyle \pm .250}$ \\
    \midrule
    ASL3DWord~\cite{SignAvatar}
    & 71.809$^{\scriptstyle \pm 2.236}$ & 0.733$^{\scriptstyle \pm .019}$ & 27.811$^{\scriptstyle \pm .430}$ & 10.483$^{\scriptstyle \pm .170}$
    & 65.037$^{\scriptstyle \pm 5.015}$ & 0.581$^{\scriptstyle \pm .029}$ & 43.715$^{\scriptstyle \pm 1.490}$ & 13.726$^{\scriptstyle \pm .315}$ \\
    MySign
    & \textbf{63.831}$^{\scriptstyle \pm 3.511}$ & \textbf{0.751}$^{\scriptstyle \pm .030}$ & \textbf{29.131}$^{\scriptstyle \pm .605}$ & \textbf{10.726}$^{\scriptstyle \pm .203}$
    & \textbf{54.166}$^{\scriptstyle \pm 2.010}$ & \textbf{0.727}$^{\scriptstyle \pm .021}$ & \textbf{48.299}$^{\scriptstyle \pm .730}$ & \textbf{22.913}$^{\scriptstyle \pm .200}$ \\
    \bottomrule
\end{tabular}
\end{table}
\section{Limitations and Future Work}
While MySign provides high-fidelity MoCap data for isolated sign generation in BIM, several limitations remain that suggest directions for future work.
First, MySign currently covers 1,000 glosses produced by five signers.
Given the linguistic specificity of BIM and the scarcity of high-quality 3D data for under-resourced sign languages, we plan to enrich its coverage and diversity with additional glosses and sign samples from the BIM Sign Bank and other BIM-specific corpora.
Second, our current data collection pipeline focuses on fine-grained body and hand kinematics and does not capture facial expressions, mouthings, or other non-manual components, which remain technically challenging to acquire in MoCap-based settings.
Incorporating these signals in subsequent releases would yield a more comprehensive representation of signing behaviors.
Third, coarticulation, sentence-level grammar, and other contextual effects lie outside the scope of our isolated-sign benchmark, so our conclusions apply to sign-internal articulation rather than continuous sign generation.
Although continuous signing is not a simple concatenation of isolated signs, MySign provides a high-fidelity lexicon of articulatory targets as a step toward it.
Finally, MoCap-grounded and RGB-fitted pseudo-label supervision have so far been compared only quantitatively (Table~\ref{tab:mocap_vs_rgb}); we plan to complement this comparison with a perceptual study of the motion generated under each type of supervision.
\section{Conclusions}
\label{sec:conclusion}

We introduce MySign, the first MoCap dataset of isolated BIM signs.
Its 5{,}000 high-fidelity samples cover 1{,}000 unique glosses and provide ground-truth SMPL-X sequences of upper-body, hand, and finger motion.
Recorded with optical MoCap for the upper body and instrumented data gloves for the hands, MySign captures fine-grained articulation with a precision rarely available in vision-based sign language datasets.
Benchmarking experiments show that MySign reliably supports supervised learning while exposing persistent gaps in state-of-the-art generation models, establishing it as a high-quality training resource and a challenging benchmark for gloss-conditioned 3D sign generation.
A controlled comparison further indicates that its MoCap-grounded motion provides more effective supervision than RGB-fitted pseudo-labels.
These results further highlight fine-grained, linguistically faithful sign generation as a key open challenge.
We anticipate that MySign will serve as a valuable resource for BIM research and lay the groundwork for advances in sign language understanding and accessibility technologies.

\begin{ack}
This work was supported in part by Monash University Malaysia through the RIE Future Center Grant Scheme (I-M010-STG-000406). We gratefully acknowledge the Monash University Malaysia Advanced Computing Platform (MUMACP) for providing the computational resources essential to this research.
\end{ack}

\bibliographystyle{plainnat}
\bibliography{reference}

\appendix

\section*{Appendix}

\section{Capture Setup}
\label{app:capture-setup}
\begin{figure} [htbp]
  \centering
  \includegraphics[width=0.85\linewidth]{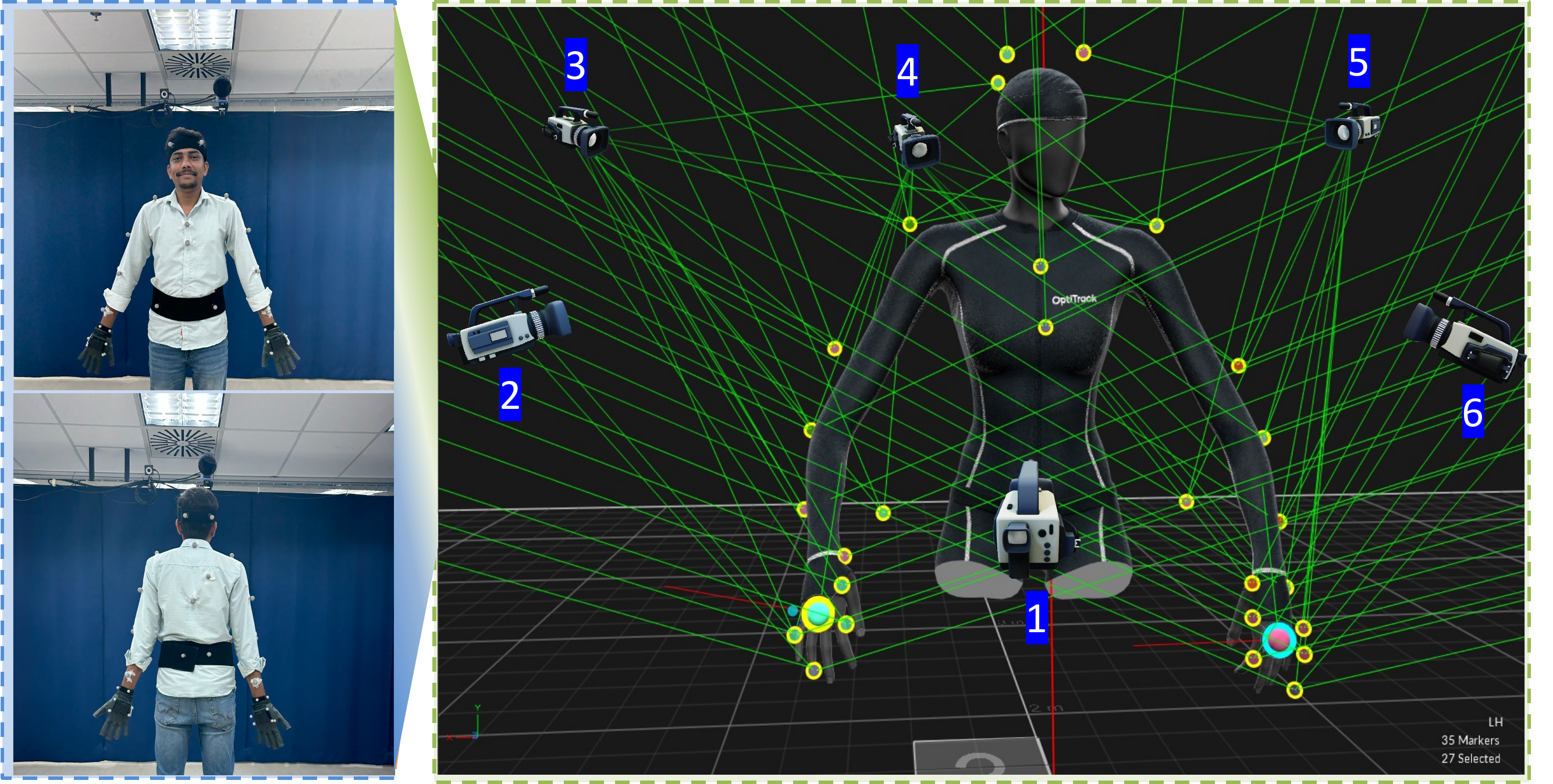}
  \caption{Dataset recording setup.}
  \label{fig:setup}
\end{figure}

Figure~\ref{fig:setup} illustrates the capture setup used for the dataset recordings.
The left panels show front and back views of a signer wearing the reflective upper-body marker set and MANUS Prime 3 data gloves.
The right panel visualizes the OptiTrack capture volume, including the six-camera arrangement, per-camera marker visibility rays, and the reconstructed 3D marker positions on the signer.
Together, these views complement the description in the main paper by showing the spatial layout of cameras and markers used during recording.

\section{Retargeting Validation, ROM-Clamping Configuration, and Clamping Ablation} 
\label{app:rom}

\subsection{Retargeting Validation}
\label{app:rom:retargeting}
Motive fuses the optical-marker and MANUS glove signals into a body-and-finger skeleton, which the Rokoko Retarget plugin transfers to SMPL-X via source-to-target bone mapping (Retarget Pose = Rest; Auto Scale = Off).
Since bone mapping, unlike model fitting, defines no fitting residual with respect to the raw signals, we instead evaluate the kinematic consistency between the fused skeleton and the pre-cleaning SMPL-X output.
On the evaluated sequences, we exclude frames flagged by Motive as gap-filled (23\% of frames) and rigidly align each remaining frame (without scaling), using the torso as the reference for the arms and the palm as the reference for each hand.
We report two metrics.
The \emph{bone-direction angle} between corresponding bones is invariant to bone length and rotation parameterization, but does not capture twist about the bone's long axis.
The \emph{normalized hand-joint position difference} is the joint position difference as a percentage of hand length; it is evaluated only on the PIP, DIP, and thumb joints, as the wrist and MCP joints define the alignment.
Table~\ref{tab:retargeting} reports the bone-direction angle per joint group; the normalized hand-joint position difference has a median of 4.5\% and a 95th percentile of 13.0\%.
Within this comparison, retargeting largely preserves bone directions and intra-hand configuration, with larger deviations at the distal finger and thumb joints.

\begin{table}[t]
\captionsetup{position=top, skip=6pt}
\caption{Retargeting validation. Bone-direction angle between the fused skeleton and the pre-cleaning SMPL-X output, reported per joint group over the evaluated sequences.}
\label{tab:retargeting}
\centering
\footnotesize
\setlength{\tabcolsep}{4pt}
\begin{tabular}{@{}lcc@{}}
\toprule
\textbf{Joint group} & \textbf{Median (\textdegree)} & \textbf{95th percentile (\textdegree)} \\
\midrule
Shoulders  & 3.0 & 8.5 \\
Elbows     & 3.5 & 9.5 \\
Wrists     & 5.0 & 13.0 \\
Finger MCP & 4.0 & 11.0 \\
Finger PIP & 5.0 & 13.0 \\
Finger DIP & 6.5 & 17.0 \\
Thumb      & 7.0 & 19.0 \\
\bottomrule
\end{tabular}
\end{table}

\subsection{Paired Clamping Ablation}
\label{app:rom:clamping}
This ablation complements the per-group statistics in Table~\ref{tab:rom_clamping} (Sec.~\ref{subsec:quality}) and focuses on the hand joints, where clipping is most frequent.
We compare each sequence without hand clamping (U) and with hand clamping (H); the two conditions are identical except for the hand-clamping step.
Table~\ref{tab:clamp_geometry} reports the resulting geometric correction over the full dataset and on HIGH.
Over the full dataset, the median correction per clipped DIP Euler component is 3.4$^\circ$.
Table~\ref{tab:clamp_semantic} reports the Top-1 accuracy and prediction agreement of the frozen ST-GCN evaluator (Sec.~\ref{subsec:recognition}) on the test set.
Since the evaluator is trained on ROM-cleaned data, H is in-distribution whereas U is not; the small Top-1 gain of H is consistent with this distribution shift and should not be read as evidence that clamping improves linguistic correctness.
Prediction agreement therefore measures lexical preservation more directly.
Within these kinematic and paired-recognition tests, clamping yields a sub-millimeter median fingertip displacement over clipped frames and leaves the evaluator's gloss predictions largely unchanged, including on HIGH.

\begin{table}[h]
  \captionsetup{position=top, skip=6pt}
  \caption{Paired clamping ablation: geometric correction. Clip rate: as in Table~\ref{tab:rom_clamping}. $|\Delta e|$: magnitude of the Euler-angle correction per clipped DIP component. Fingertip displacement: computed with wrist-anchored hand-local forward kinematics. HIGH: top 10\% of test sequences by per-sequence P95 fingertip displacement ($N = 100$), fixed before any recognition output was inspected.}
  \label{tab:clamp_geometry}
  \centering
  \footnotesize
  \setlength{\tabcolsep}{6pt}
  \begin{tabular}{lccccccc}
    \toprule
    \multirow{2}{*}{\textbf{Scope}} & \multirow{2}{*}{\textbf{DIP clip rate (\%)}} & \multicolumn{3}{c}{\textbf{$|\Delta e|$ (\textdegree)}} & \multicolumn{3}{c}{\textbf{Fingertip displacement (mm)}} \\
    \cmidrule(lr){3-5} \cmidrule(lr){6-8}
     & & \textbf{Median} & \textbf{P95} & \textbf{Max} & \textbf{Median} & \textbf{P95} & \textbf{Max} \\
    \midrule
    Full dataset & 21.04 & 3.4 & 16.5 & 82 & 0.15 & 3.1 & 28.1 \\
    HIGH         & 46    & 5.2 & 25.3 & 62 & 1.9  & 8.4 & 26   \\
    \bottomrule
  \end{tabular}
\end{table}

\begin{table}[h]
  \captionsetup{position=top, skip=6pt}
  \caption{Paired clamping ablation: recognition with the frozen ST-GCN evaluator (Sec.~\ref{subsec:recognition}). U: without hand clamping; H: hand-clamped; $\Delta$ = H $-$ U. Top-1 is mean $\pm$ 95\% CI over 20 evaluation repetitions. Agreement: share of sequences for which U and H receive the same predicted gloss. HIGH as in Table~\ref{tab:clamp_geometry}.}
  \label{tab:clamp_semantic}
  \centering
  \footnotesize
  \setlength{\tabcolsep}{6pt}
\begin{tabular}{lccccc}
  \toprule
  \textbf{Scope} & \textbf{$N$} &
  \textbf{U Top-1 (\%)} & \textbf{H Top-1 (\%)} &
  \textbf{$\Delta$ (pp)} & \textbf{Pred. agreement (\%)} \\
  \midrule
  All test & 1,000 & $85.0 \pm 0.3$ & $85.7 \pm 0.2$ & $+0.5$ & 96.8 \\
  HIGH     &   100 & $82.0 \pm 0.7$ & $83.0 \pm 0.5$ & $+1.0$ & 92.0 \\
  \bottomrule
\end{tabular}
\end{table}

\subsection{ROM-Clamping Configuration}
\label{app:rom:config}

The released configuration specifies per-joint Euler bounds $[\mathbf{e}_j^{\min}, \mathbf{e}_j^{\max}]$ for the upper-limb and finger joints and per-joint change-of-basis transforms $\mathbf{B}_j$ for the finger joints, with separate entries for the left and right hands whose values are numerical mirror images of each other.
Upper-limb bounds are applied in the native SMPL-X joint frames, whose axes are sufficiently aligned with the anatomical axes, whereas finger bounds are applied in the anatomical frames defined by $\mathbf{B}_j$.
In both cases, clamping is component-wise in an Euler chart, which approximates exact projection onto the ROM region and matches how anatomical limits are specified, namely along flexion, abduction, and rotation axes.
MCP joints admit flexion and abduction but no axial rotation, and PIP and DIP joints admit flexion only.

\section{Expert-Based Perceptual Validation: Protocol Details}
\label{app:verify}
We provide additional details on the expert-based perceptual validation procedure summarized in the main paper.
We describe the rater workflow through the web interface, the rating scale used for motion naturalness, and the criterion used to determine whether a perceived meaning matches the reference gloss.

\subsection{Rater Workflow}
\label{app:verify:workflow}
Three Deaf native signers with BIM domain expertise independently completed the validation through a dedicated web interface.
The workflow proceeded as follows:

\begin{enumerate}[leftmargin=*, nosep]
    \item \textbf{Login.} The expert opens the validation website and enters their name; this identifier is recorded with all subsequent ratings to enable inter-rater analysis.

    \item \textbf{Session structure.} Sign instances are organized into sessions of 200 signs each, allowing experts to complete the validation
    over multiple sittings. Within a session, signs are presented one at a time.

    \item \textbf{Per-sign procedure.} For each sign, the expert clicks \emph{play} to view a video rendering of the 3D motion. The reference gloss is not displayed at any point, so the expert sees only the rendered motion. After the video completes, the expert:
    \begin{enumerate}[leftmargin=*, nosep]
        \item Enters the perceived meaning of the sign as a free-text response in English, and
        \item Rates the naturalness of the motion on the five-level scale described in Appendix~\ref{app:verify:anchors}.
    \end{enumerate}

    \item \textbf{Advancing.} After both fields are filled in, the expert clicks \emph{next} to record the response and load the next sign. Signs cannot be skipped, and responses are auto-saved to allow sessions to be resumed.
\end{enumerate}

\begin{figure}[t!]
  \centering
  \includegraphics[width=0.85\linewidth]{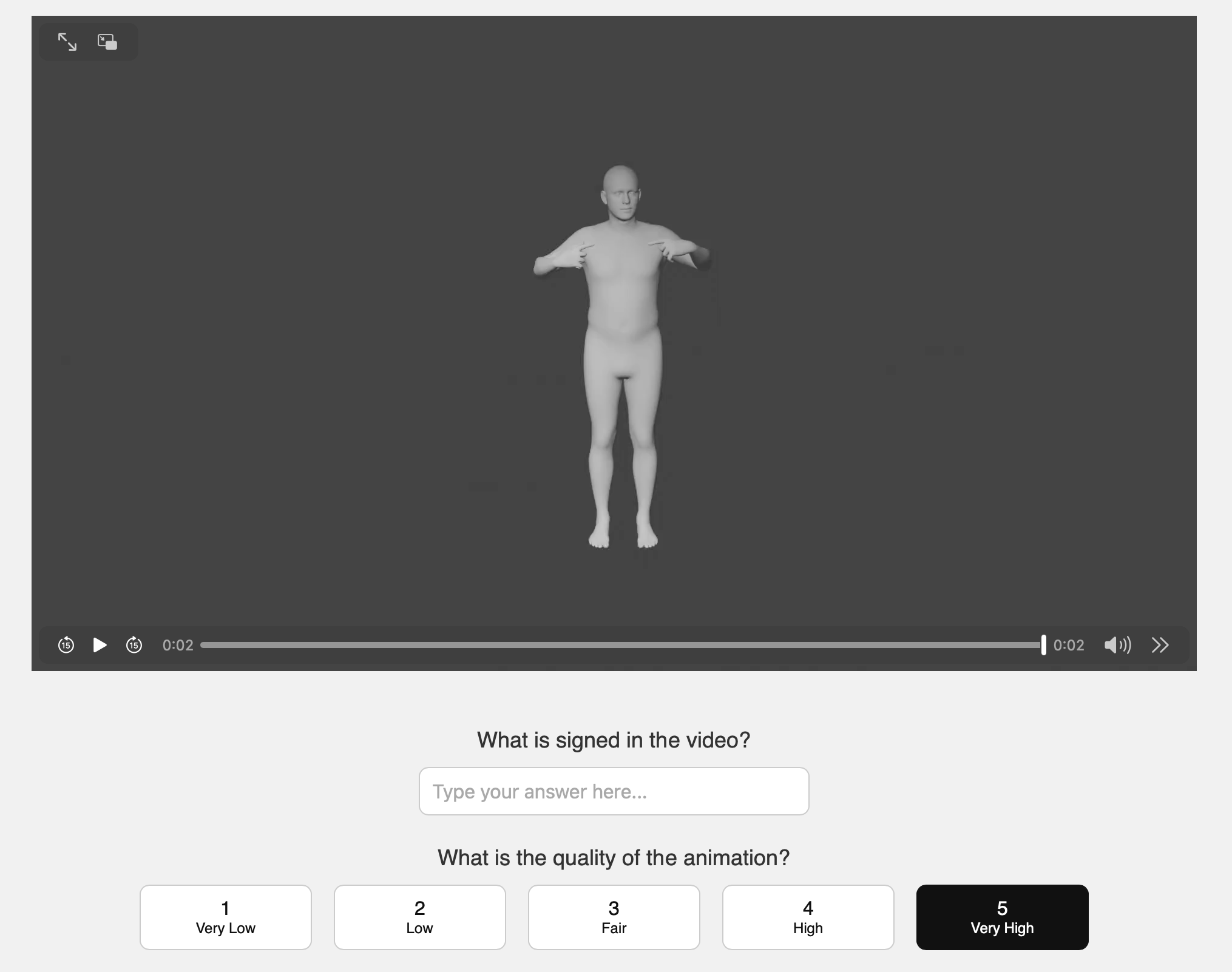}
  \caption{Web interface used by domain experts for the perceptual validation study.}
  \label{fig:verify_ui}
\end{figure}

A screenshot of the interface is shown in Figure~\ref{fig:verify_ui}.

\subsection{Naturalness Rating Anchors}
\label{app:verify:anchors}
Experts rated the naturalness of each rendered motion on a five-level Likert scale.
To reduce inter-rater variability in scale interpretation, each level was accompanied by an anchor description shown alongside the rating buttons in the interface:

\begin{itemize}[leftmargin=*, nosep]
    \item \textbf{Very high (5).} The motion is fluent and natural, with smooth trajectories and well-coordinated hand and finger articulation; no visible artifacts.
    \item \textbf{High (4).} The motion is largely natural and clearly conveys the sign; minor imperfections may be present but do not distract from the overall articulation.
    \item \textbf{Fair (3).} The motion is acceptable and the sign remains discernible, but noticeable imperfections are present, such as slightly stiff transitions or imprecise finger configurations.
    \item \textbf{Low (2).} The motion shows clear unnaturalness, with visible artifacts in trajectory or hand shape that interfere with fluent articulation, although the sign may still be partially discernible.
    \item \textbf{Very low (1).} The motion is severely unnatural, with pronounced artifacts (e.g., implausible joint configurations, jittery or broken motion) that substantially degrade the rendered sign.
\end{itemize}

Experts were instructed to rate naturalness based on the rendered SMPL-X sequence as observed in the interface, independently of whether the sign's meaning was correctly identified.

\subsection{Gloss-Matching Criterion}
\label{app:verify:matching}
To compute the recognizability statistic reported in the main paper, each expert's free-text English response was compared against the reference gloss using a strict lexical-match criterion.
A response was counted as matching the reference gloss when the two strings were lexically identical after the following normalization steps applied to both:

\begin{itemize}[leftmargin=*, nosep]
    \item Case folding (case-insensitive comparison).
    \item Trimming of leading and trailing whitespace, and collapsing of repeated internal whitespace.
    \item Removal of hyphens and equivalent treatment of single-word and hyphenated forms (e.g., \emph{ice cream}, \emph{ice-cream}, and \emph{icecream} were treated as equivalent).
    \item Lemmatization to a common base form, so that inflectional variants such as plural forms (e.g., \emph{cats} vs.\ \emph{cat}) and standard verb inflections (e.g., \emph{running}, \emph{ran} vs.\ \emph{run}) were treated as equivalent.
\end{itemize}

Beyond the above normalization, no further relaxation was applied: synonyms, hypernyms, hyponyms, paraphrases, semantically related terms, and non-English responses were \emph{not} counted as matches.
We adopted this strict criterion to provide a conservative lower bound on perceptual recognizability; the figure of $86.3\%$ reported in the main paper should therefore be interpreted as a lower-bound estimate, since responses that were semantically correct but lexically distinct from the reference gloss were excluded from the match count.

\section{Hand-Motion Characterization}
\label{app:motion_visualization}
This appendix provides supporting visualizations and analysis for the articulation and hand-pose variation properties referenced in the main paper.

\begin{figure}[h]
  \centering
  \includegraphics[width=0.85\linewidth]{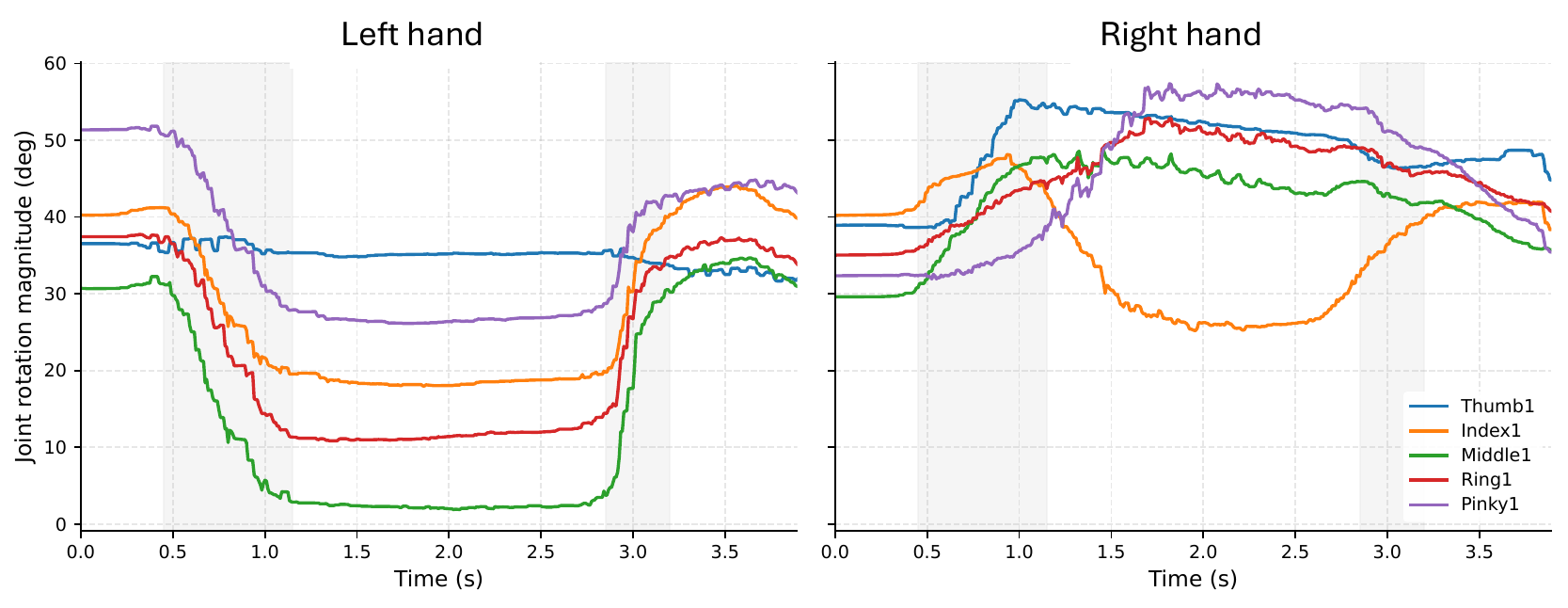}
  \caption{Finger joint articulation trajectories for a representative sign sequence. Gray bands mark major handshape transition intervals.}
  \label{fig:articulation_trajectories}
\end{figure}

\begin{figure}[h]
  \centering
  \includegraphics[width=0.75\linewidth]{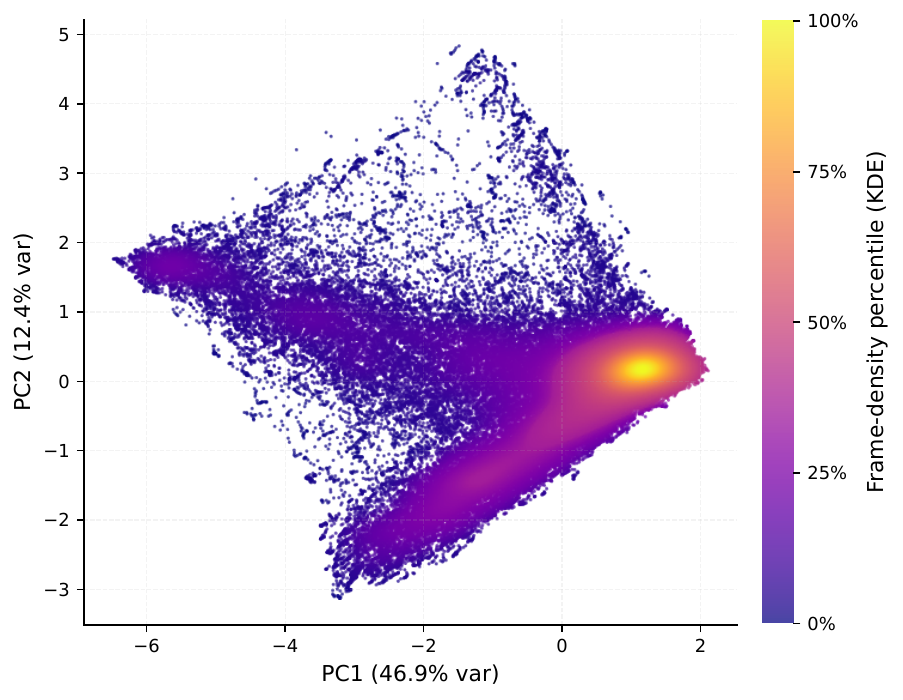}
  \caption{PCA projection of bimanual hand poses. Color denotes the percentile rank of per-frame kernel density in the 2D projection.}
  \label{fig:bimanual_handpose_variation}
\end{figure}

\noindent\textbf{Finger joint articulation trajectories.}
Figure~\ref{fig:articulation_trajectories} plots per-frame SMPL-X joint rotation magnitudes (in degrees) for the proximal joints of all five fingers on both the left and right hands, taken from a representative sign sequence captured at 200\,Hz.
Each curve corresponds to a single proximal finger joint, and the left- and right-hand panels are shown separately to expose bimanual structure.
Gray vertical bands mark major handshape transition intervals, identified as windows of high aggregate angular velocity across the depicted joints.
Three properties are visible in this representation:
(i)~The trajectories evolve smoothly across frames, indicating temporal continuity of the captured articulation; (ii)~Individual fingers follow distinct trajectories rather than moving as a single rigid unit, which we refer to as finger-specific articulation; and (iii)~The left- and right-hand trajectories are not mirror images of one another, reflecting the bimanual asymmetry that is characteristic of natural BIM production.
Together, these properties indicate that the capture resolves fine-grained articulatory structure rather than approximating each hand as a rigid segment.

\noindent\textbf{Bimanual hand-pose variation (PCA).}
Figure~\ref{fig:bimanual_handpose_variation} visualizes the distribution of bimanual hand poses in a low-dimensional projection.
For each sampled frame, we extract the 30 SMPL-X hand-joint axis-angle rotations (15 per hand) and convert each rotation to the 6D continuous rotation representation of Zhou~\emph{et al.}~\cite{6Dcontinuity}, yielding a 180-D per-frame feature.
We then apply PCA to the resulting feature matrix; the first two principal components explain 46.9\% and 12.4\% of the total variance, respectively.
Each point in the scatter plot corresponds to a single sampled frame projected onto PC1 and PC2.
Color encodes the percentile rank of the per-frame Gaussian kernel density estimate (KDE) computed in the 2D projection, so the color field reflects frame occupancy across the sampled frames rather than handshape frequency in any phonological inventory.
The projection occupies a broad, multi-modal region of the PC1--PC2 plane and exhibits an extended low-density tail, which we interpret as qualitative visual evidence consistent with substantial hand-pose variation across the dataset.

\noindent\textbf{Scope of the claim.}
Figure~\ref{fig:bimanual_handpose_variation} is intended as a qualitative illustration of hand-pose variation rather than a standalone measure of dataset superiority, BIM handshape coverage, or pose fidelity.
The first two principal components together account for approximately 59\% of the variance in the 180-D feature space, so the projection necessarily compresses higher-order structure, and the KDE coloring describes the empirical density of sampled frames rather than the linguistic distribution of BIM handshapes.
Claims regarding coverage, fidelity, and comparative advantage are addressed in the experimental sections of the paper.

\section{Experimental Settings}
\label{app:experimental_settings}

All models used in this work are publicly available and linked below.

\begin{itemize}[leftmargin=*, nosep]
    \item \textbf{Isolated sign language generation:} 
    SignAvatar~\cite{SignAvatar}~\href{https://github.com/dongludeeplearning/SignAvatar}{\faGithub};
    wSignGen~\cite{wsigngen}~\href{https://github.com/dongludeeplearning/wsigngen}{\faGithub}.

    \item \textbf{Action-conditioned human motion generation:} 
    ACTOR~\cite{actor}~\href{https://github.com/Mathux/ACTOR}{\faGithub};
    MDM~\cite{mdm}~\href{https://guytevet.github.io/mdm-page/}{\faGithub};
    MLD~\cite{mld}~\href{https://github.com/chenfengye/motion-latent-diffusion}{\faGithub};
    MotionDiffuse~\cite{motiondiffuse}~\href{https://github.com/MotrixLab/MotionDiffuse}{\faGithub};
    LS-GAN~\cite{Ls-gan}~\href{https://github.com/AmballaAvinash/motion-latent-diffusion}{\faGithub}.

    \item \textbf{Isolated sign language recognition:} 
    ST-GCN~\cite{stgcn}~\href{https://github.com/itskalvik/ST-GCN}{\faGithub};
    SignBERT~\cite{signbert}~\href{https://github.com/joshuasv/signbert_unofficial}{\faGithub};
    SignBERT+~\cite{signbert+}~\href{https://github.com/joshuasv/signbert_unofficial}{\faGithub}.
\end{itemize}

We express our profound gratitude to the aforementioned authors for their invaluable contributions.

Our experiments are implemented in PyTorch~\cite{PyTorch} and models are trained on NVIDIA A100 GPUs with 40\,GB memory.
We use the default hyperparameters for training the models of gloss-conditioned isolated sign generation and sign language recognition.

\section{Additional Experimental Results}
\label{app:additional}

\noindent\textbf{Samples-per-gloss scaling.}
Since each signer records each gloss once (Sec.~\ref{subsec:splits}), the number of samples per gloss equals the number of signers per gloss, $n$; since the predefined split holds out one signer, the range available for training is $n \in \{1,\dots,4\}$.
For each $n$, we train wSignGen~\cite{wsigngen} on every subset of $n$ of the four training signers (15 subsets in total), keeping the held-out signer and the training and evaluation configurations fixed and using the same frozen motion encoder and ST-GCN evaluator (Sec.~\ref{subsec:recognition}) for every $n$.
As Table~\ref{tab:scaling} shows, FID consistently decreases and Acc consistently increases with $n$, while Div and Multimod move toward the Real reference.
We therefore read the ablation as evidence that broader inter-signer coverage is beneficial, without inferring saturation or sample sufficiency at four or five samples per gloss.

\begin{table}[h]
  \captionsetup{position=top, skip=6pt}
  \caption{Samples-per-gloss scaling ablation with wSignGen. $n$ denotes the number of signers per gloss used for training; for $n < 4$, results are averaged over all subsets of the four training signers (4, 6, and 4 subsets). All values, including the Real row, are mean $\pm$ 95\% CI over 20 repetitions.}
  \label{tab:scaling}
  \centering
  \footnotesize
  \setlength{\tabcolsep}{6pt}
  \begin{tabular}{ccccccc}
    \toprule
    \textbf{$n$} & \textbf{Train seq.} & \textbf{\# subsets} & \textbf{FID$\downarrow$} & \textbf{Acc$\uparrow$} & \textbf{Div$\rightarrow$} & \textbf{Multimod$\rightarrow$} \\
    \midrule
    Real & -- & -- & 0.023$^{\scriptstyle \pm .020}$ & 0.857$^{\scriptstyle \pm .002}$ & 46.881$^{\scriptstyle \pm .300}$ & 18.719$^{\scriptstyle \pm .250}$ \\
    \midrule
    1 & 1{,}000 & 4 & 123.503$^{\scriptstyle \pm 6.544}$ & 0.500$^{\scriptstyle \pm .021}$ & 55.508$^{\scriptstyle \pm 1.300}$ & 30.214$^{\scriptstyle \pm .463}$ \\
    2 & 2{,}000 & 6 & 82.549$^{\scriptstyle \pm 4.416}$ & 0.629$^{\scriptstyle \pm .017}$ & 52.442$^{\scriptstyle \pm 1.054}$ & 26.520$^{\scriptstyle \pm .382}$ \\
    3 & 3{,}000 & 4 & 58.536$^{\scriptstyle \pm 3.035}$ & 0.705$^{\scriptstyle \pm .014}$ & 50.739$^{\scriptstyle \pm .904}$ & 24.426$^{\scriptstyle \pm .314}$ \\
    4 & 4{,}000 & 1 & 41.465$^{\scriptstyle \pm 2.069}$ & 0.758$^{\scriptstyle \pm .013}$ & 49.407$^{\scriptstyle \pm .810}$ & 22.825$^{\scriptstyle \pm .230}$ \\
    \bottomrule
  \end{tabular}
\end{table}

\newpage
\section*{NeurIPS Paper Checklist}

\begin{enumerate}

\item {\bf Claims}
    \item[] Question: Do the main claims made in the abstract and introduction accurately reflect the paper's contributions and scope?
    \item[] Answer: \answerYes{} 
    \item[] Justification: We discuss the motivation and contribution of the dataset in these sections.
    \item[] Guidelines:
    \begin{itemize}
        \item The answer \answerNA{} means that the abstract and introduction do not include the claims made in the paper.
        \item The abstract and/or introduction should clearly state the claims made, including the contributions made in the paper and important assumptions and limitations. A \answerNo{} or \answerNA{} answer to this question will not be perceived well by the reviewers. 
        \item The claims made should match theoretical and experimental results, and reflect how much the results can be expected to generalize to other settings. 
        \item It is fine to include aspirational goals as motivation as long as it is clear that these goals are not attained by the paper. 
    \end{itemize}

\item {\bf Limitations}
    \item[] Question: Does the paper discuss the limitations of the work performed by the authors?
    \item[] Answer: \answerYes{} 
    \item[] Justification: The paper includes a dedicated ``Limitations and Future Work'' section that explicitly discusses limitations.
    \item[] Guidelines:
    \begin{itemize}
        \item The answer \answerNA{} means that the paper has no limitation while the answer \answerNo{} means that the paper has limitations, but those are not discussed in the paper. 
        \item The authors are encouraged to create a separate ``Limitations'' section in their paper.
        \item The paper should point out any strong assumptions and how robust the results are to violations of these assumptions (e.g., independence assumptions, noiseless settings, model well-specification, asymptotic approximations only holding locally). The authors should reflect on how these assumptions might be violated in practice and what the implications would be.
        \item The authors should reflect on the scope of the claims made, e.g., if the approach was only tested on a few datasets or with a few runs. In general, empirical results often depend on implicit assumptions, which should be articulated.
        \item The authors should reflect on the factors that influence the performance of the approach. For example, a facial recognition algorithm may perform poorly when image resolution is low or images are taken in low lighting. Or a speech-to-text system might not be used reliably to provide closed captions for online lectures because it fails to handle technical jargon.
        \item The authors should discuss the computational efficiency of the proposed algorithms and how they scale with dataset size.
        \item If applicable, the authors should discuss possible limitations of their approach to address problems of privacy and fairness.
        \item While the authors might fear that complete honesty about limitations might be used by reviewers as grounds for rejection, a worse outcome might be that reviewers discover limitations that aren't acknowledged in the paper. The authors should use their best judgment and recognize that individual actions in favor of transparency play an important role in developing norms that preserve the integrity of the community. Reviewers will be specifically instructed to not penalize honesty concerning limitations.
    \end{itemize}

\item {\bf Theory assumptions and proofs}
    \item[] Question: For each theoretical result, does the paper provide the full set of assumptions and a complete (and correct) proof?
    \item[] Answer: \answerNA{} 
    \item[] Justification: The paper does not include theoretical results.
    \item[] Guidelines:
    \begin{itemize}
        \item The answer \answerNA{} means that the paper does not include theoretical results. 
        \item All the theorems, formulas, and proofs in the paper should be numbered and cross-referenced.
        \item All assumptions should be clearly stated or referenced in the statement of any theorems.
        \item The proofs can either appear in the main paper or the supplemental material, but if they appear in the supplemental material, the authors are encouraged to provide a short proof sketch to provide intuition. 
        \item Inversely, any informal proof provided in the core of the paper should be complemented by formal proofs provided in appendix or supplemental material.
        \item Theorems and Lemmas that the proof relies upon should be properly referenced. 
    \end{itemize}

    \item {\bf Experimental result reproducibility}
    \item[] Question: Does the paper fully disclose all the information needed to reproduce the main experimental results of the paper to the extent that it affects the main claims and/or conclusions of the paper (regardless of whether the code and data are provided or not)?
    \item[] Answer: \answerYes{} 
    \item[] Justification: Benchmark methodology and evaluation metrics are detailed in Sec.~\ref{Sec:exp} and Appendix~\ref{app:experimental_settings}.
    \item[] Guidelines:
    \begin{itemize}
        \item The answer \answerNA{} means that the paper does not include experiments.
        \item If the paper includes experiments, a \answerNo{} answer to this question will not be perceived well by the reviewers: Making the paper reproducible is important, regardless of whether the code and data are provided or not.
        \item If the contribution is a dataset and\slash or model, the authors should describe the steps taken to make their results reproducible or verifiable. 
        \item Depending on the contribution, reproducibility can be accomplished in various ways. For example, if the contribution is a novel architecture, describing the architecture fully might suffice, or if the contribution is a specific model and empirical evaluation, it may be necessary to either make it possible for others to replicate the model with the same dataset, or provide access to the model. In general. releasing code and data is often one good way to accomplish this, but reproducibility can also be provided via detailed instructions for how to replicate the results, access to a hosted model (e.g., in the case of a large language model), releasing of a model checkpoint, or other means that are appropriate to the research performed.
        \item While NeurIPS does not require releasing code, the conference does require all submissions to provide some reasonable avenue for reproducibility, which may depend on the nature of the contribution. For example
        \begin{enumerate}
            \item If the contribution is primarily a new algorithm, the paper should make it clear how to reproduce that algorithm.
            \item If the contribution is primarily a new model architecture, the paper should describe the architecture clearly and fully.
            \item If the contribution is a new model (e.g., a large language model), then there should either be a way to access this model for reproducing the results or a way to reproduce the model (e.g., with an open-source dataset or instructions for how to construct the dataset).
            \item We recognize that reproducibility may be tricky in some cases, in which case authors are welcome to describe the particular way they provide for reproducibility. In the case of closed-source models, it may be that access to the model is limited in some way (e.g., to registered users), but it should be possible for other researchers to have some path to reproducing or verifying the results.
        \end{enumerate}
    \end{itemize}

\item {\bf Open access to data and code}
    \item[] Question: Does the paper provide open access to the data and code, with sufficient instructions to faithfully reproduce the main experimental results, as described in supplemental material?
    \item[] Answer: \answerYes{} 
    \item[] Justification: Our datasets are hosted on Hugging Face:
    \url{https://huggingface.co/datasets/mysigner/MySign}
    \item[] Guidelines:
    \begin{itemize}
        \item The answer \answerNA{} means that paper does not include experiments requiring code.
        \item Please see the NeurIPS code and data submission guidelines (\url{https://neurips.cc/public/guides/CodeSubmissionPolicy}) for more details.
        \item While we encourage the release of code and data, we understand that this might not be possible, so \answerNo{} is an acceptable answer. Papers cannot be rejected simply for not including code, unless this is central to the contribution (e.g., for a new open-source benchmark).
        \item The instructions should contain the exact command and environment needed to run to reproduce the results. See the NeurIPS code and data submission guidelines (\url{https://neurips.cc/public/guides/CodeSubmissionPolicy}) for more details.
        \item The authors should provide instructions on data access and preparation, including how to access the raw data, preprocessed data, intermediate data, and generated data, etc.
        \item The authors should provide scripts to reproduce all experimental results for the new proposed method and baselines. If only a subset of experiments are reproducible, they should state which ones are omitted from the script and why.
        \item At submission time, to preserve anonymity, the authors should release anonymized versions (if applicable).
        \item Providing as much information as possible in supplemental material (appended to the paper) is recommended, but including URLs to data and code is permitted.
    \end{itemize}

\item {\bf Experimental setting/details}
    \item[] Question: Does the paper specify all the training and test details (e.g., data splits, hyperparameters, how they were chosen, type of optimizer) necessary to understand the results?
    \item[] Answer: \answerYes{} 
    \item[] Justification: Experimental settings are described in Sec.~\ref{Sec:exp} and Appendix~\ref{app:experimental_settings}.
    \item[] Guidelines:
    \begin{itemize}
        \item The answer \answerNA{} means that the paper does not include experiments.
        \item The experimental setting should be presented in the core of the paper to a level of detail that is necessary to appreciate the results and make sense of them.
        \item The full details can be provided either with the code, in appendix, or as supplemental material.
    \end{itemize}

\item {\bf Experiment statistical significance}
    \item[] Question: Does the paper report error bars suitably and correctly defined or other appropriate information about the statistical significance of the experiments?
    \item[] Answer: \answerYes{} 
    \item[] Justification: The main experimental results are reported with 95\% confidence intervals, with the type of interval stated in the surrounding text rather than the table captions; the exact number of training seeds underlying these intervals is not separately specified in the paper.
    \item[] Guidelines:
    \begin{itemize}
        \item The answer \answerNA{} means that the paper does not include experiments.
        \item The authors should answer \answerYes{} if the results are accompanied by error bars, confidence intervals, or statistical significance tests, at least for the experiments that support the main claims of the paper.
        \item The factors of variability that the error bars are capturing should be clearly stated (for example, train/test split, initialization, random drawing of some parameter, or overall run with given experimental conditions).
        \item The method for calculating the error bars should be explained (closed form formula, call to a library function, bootstrap, etc.)
        \item The assumptions made should be given (e.g., Normally distributed errors).
        \item It should be clear whether the error bar is the standard deviation or the standard error of the mean.
        \item It is OK to report 1-sigma error bars, but one should state it. The authors should preferably report a 2-sigma error bar than state that they have a 96\% CI, if the hypothesis of Normality of errors is not verified.
        \item For asymmetric distributions, the authors should be careful not to show in tables or figures symmetric error bars that would yield results that are out of range (e.g., negative error rates).
        \item If error bars are reported in tables or plots, the authors should explain in the text how they were calculated and reference the corresponding figures or tables in the text.
    \end{itemize}

\item {\bf Experiments compute resources}
    \item[] Question: For each experiment, does the paper provide sufficient information on the computer resources (type of compute workers, memory, time of execution) needed to reproduce the experiments?
    \item[] Answer: \answerYes{} 
    \item[] Justification: The experiments compute resources are described in Appendix~\ref{app:experimental_settings}.
    \item[] Guidelines:
    \begin{itemize}
        \item The answer \answerNA{} means that the paper does not include experiments.
        \item The paper should indicate the type of compute workers CPU or GPU, internal cluster, or cloud provider, including relevant memory and storage.
        \item The paper should provide the amount of compute required for each of the individual experimental runs as well as estimate the total compute. 
        \item The paper should disclose whether the full research project required more compute than the experiments reported in the paper (e.g., preliminary or failed experiments that didn't make it into the paper). 
    \end{itemize}
    
\item {\bf Code of ethics}
    \item[] Question: Does the research conducted in the paper conform, in every respect, with the NeurIPS Code of Ethics \url{https://neurips.cc/public/EthicsGuidelines}?
    \item[] Answer: \answerYes{} 
    \item[] Justification: The research fully complies with the NeurIPS Code of Ethics.
    \item[] Guidelines:
    \begin{itemize}
        \item The answer \answerNA{} means that the authors have not reviewed the NeurIPS Code of Ethics.
        \item If the authors answer \answerNo, they should explain the special circumstances that require a deviation from the Code of Ethics.
        \item The authors should make sure to preserve anonymity (e.g., if there is a special consideration due to laws or regulations in their jurisdiction).
    \end{itemize}

\item {\bf Broader impacts}
    \item[] Question: Does the paper discuss both potential positive societal impacts and negative societal impacts of the work performed?
    \item[] Answer: \answerYes{} 
    \item[] Justification: We addressed this in the final section (Sec.~\ref{sec:conclusion}). In brief, we do not think that there are negative implications, but many potentially positive ones, as MySign enables characterizing fundamental sign language motion behaviors, which is useful for assessing and improving sign language understanding, as well as advancing accessibility technologies.
    \item[] Guidelines:
    \begin{itemize}
        \item The answer \answerNA{} means that there is no societal impact of the work performed.
        \item If the authors answer \answerNA{} or \answerNo, they should explain why their work has no societal impact or why the paper does not address societal impact.
        \item Examples of negative societal impacts include potential malicious or unintended uses (e.g., disinformation, generating fake profiles, surveillance), fairness considerations (e.g., deployment of technologies that could make decisions that unfairly impact specific groups), privacy considerations, and security considerations.
        \item The conference expects that many papers will be foundational research and not tied to particular applications, let alone deployments. However, if there is a direct path to any negative applications, the authors should point it out. For example, it is legitimate to point out that an improvement in the quality of generative models could be used to generate Deepfakes for disinformation. On the other hand, it is not needed to point out that a generic algorithm for optimizing neural networks could enable people to train models that generate Deepfakes faster.
        \item The authors should consider possible harms that could arise when the technology is being used as intended and functioning correctly, harms that could arise when the technology is being used as intended but gives incorrect results, and harms following from (intentional or unintentional) misuse of the technology.
        \item If there are negative societal impacts, the authors could also discuss possible mitigation strategies (e.g., gated release of models, providing defenses in addition to attacks, mechanisms for monitoring misuse, mechanisms to monitor how a system learns from feedback over time, improving the efficiency and accessibility of ML).
    \end{itemize}
    
\item {\bf Safeguards}
    \item[] Question: Does the paper describe safeguards that have been put in place for responsible release of data or models that have a high risk for misuse (e.g., pre-trained language models, image generators, or scraped datasets)?
    \item[] Answer: \answerYes{} 
    \item[] Justification: We will release our dataset under a permissive license for academic research, likely CC BY-NC-SA 4.0, upon acceptance. Following our institution's ethics guidelines, participant data have been anonymized, and all participants provided informed consent and were remunerated.
    \item[] Guidelines:
    \begin{itemize}
        \item The answer \answerNA{} means that the paper poses no such risks.
        \item Released models that have a high risk for misuse or dual-use should be released with necessary safeguards to allow for controlled use of the model, for example by requiring that users adhere to usage guidelines or restrictions to access the model or implementing safety filters. 
        \item Datasets that have been scraped from the Internet could pose safety risks. The authors should describe how they avoided releasing unsafe images.
        \item We recognize that providing effective safeguards is challenging, and many papers do not require this, but we encourage authors to take this into account and make a best faith effort.
    \end{itemize}

\item {\bf Licenses for existing assets}
    \item[] Question: Are the creators or original owners of assets (e.g., code, data, models), used in the paper, properly credited and are the license and terms of use explicitly mentioned and properly respected?
    \item[] Answer: \answerYes{} 
    \item[] Justification: All models and datasets that we used are cited and credited.
    \item[] Guidelines:
    \begin{itemize}
        \item The answer \answerNA{} means that the paper does not use existing assets.
        \item The authors should cite the original paper that produced the code package or dataset.
        \item The authors should state which version of the asset is used and, if possible, include a URL.
        \item The name of the license (e.g., CC-BY 4.0) should be included for each asset.
        \item For scraped data from a particular source (e.g., website), the copyright and terms of service of that source should be provided.
        \item If assets are released, the license, copyright information, and terms of use in the package should be provided. For popular datasets, \url{paperswithcode.com/datasets} has curated licenses for some datasets. Their licensing guide can help determine the license of a dataset.
        \item For existing datasets that are re-packaged, both the original license and the license of the derived asset (if it has changed) should be provided.
        \item If this information is not available online, the authors are encouraged to reach out to the asset's creators.
    \end{itemize}

\item {\bf New assets}
    \item[] Question: Are new assets introduced in the paper well documented and is the documentation provided alongside the assets?
    \item[] Answer: \answerYes{} 
    \item[] Justification: The paper thoroughly documents the newly introduced MySign dataset, including its structure, post-processing details, and data cleaning pipeline.
    \item[] Guidelines:
    \begin{itemize}
        \item The answer \answerNA{} means that the paper does not release new assets.
        \item Researchers should communicate the details of the dataset\slash code\slash model as part of their submissions via structured templates. This includes details about training, license, limitations, etc. 
        \item The paper should discuss whether and how consent was obtained from people whose asset is used.
        \item At submission time, remember to anonymize your assets (if applicable). You can either create an anonymized URL or include an anonymized zip file.
    \end{itemize}

\item {\bf Crowdsourcing and research with human subjects}
    \item[] Question: For crowdsourcing experiments and research with human subjects, does the paper include the full text of instructions given to participants and screenshots, if applicable, as well as details about compensation (if any)? 
    \item[] Answer: \answerYes{} 
    \item[] Justification: The paper involves research with human subjects, and the full instructions given to participants, relevant screenshots, and compensation details are included in the Sec.~\ref{sec:mysign}.
    \item[] Guidelines:
    \begin{itemize}
        \item The answer \answerNA{} means that the paper does not involve crowdsourcing nor research with human subjects.
        \item Including this information in the supplemental material is fine, but if the main contribution of the paper involves human subjects, then as much detail as possible should be included in the main paper. 
        \item According to the NeurIPS Code of Ethics, workers involved in data collection, curation, or other labor should be paid at least the minimum wage in the country of the data collector. 
    \end{itemize}

\item {\bf Institutional review board (IRB) approvals or equivalent for research with human subjects}
    \item[] Question: Does the paper describe potential risks incurred by study participants, whether such risks were disclosed to the subjects, and whether Institutional Review Board (IRB) approvals (or an equivalent approval/review based on the requirements of your country or institution) were obtained?
    \item[] Answer: \answerYes{} 
    \item[] Justification: The data collection protocol involving human subjects was reviewed and approved by the IRB at our institution. All participants were informed about the purpose of the study and any potential risks before giving their consent.
    \item[] Guidelines:
    \begin{itemize}
        \item The answer \answerNA{} means that the paper does not involve crowdsourcing nor research with human subjects.
        \item Depending on the country in which research is conducted, IRB approval (or equivalent) may be required for any human subjects research. If you obtained IRB approval, you should clearly state this in the paper. 
        \item We recognize that the procedures for this may vary significantly between institutions and locations, and we expect authors to adhere to the NeurIPS Code of Ethics and the guidelines for their institution. 
        \item For initial submissions, do not include any information that would break anonymity (if applicable), such as the institution conducting the review.
    \end{itemize}

\item {\bf Declaration of LLM usage}
    \item[] Question: Does the paper describe the usage of LLMs if it is an important, original, or non-standard component of the core methods in this research? Note that if the LLM is used only for writing, editing, or formatting purposes and does \emph{not} impact the core methodology, scientific rigor, or originality of the research, declaration is not required.
    \item[] Answer: \answerNA{} 
    \item[] Justification: The core method development in this research does not involve LLMs.
    \item[] Guidelines:
    \begin{itemize}
        \item The answer \answerNA{} means that the core method development in this research does not involve LLMs as any important, original, or non-standard components.
        \item Please refer to our LLM policy in the NeurIPS handbook for what should or should not be described.
    \end{itemize}

\end{enumerate}

\end{document}